\documentclass{article}

     \usepackage[preprint]{neurips_2022}

\usepackage[utf8]{inputenc} 
\usepackage[T1]{fontenc}    
\usepackage{hyperref}       
\usepackage{url}            
\usepackage{booktabs}       
\usepackage{amsfonts}       
\usepackage{nicefrac}       
\usepackage{microtype}      
\usepackage{xcolor}         
\usepackage{graphicx}
\usepackage{comment}
\usepackage{amsmath,amssymb}
\usepackage{listings}

\usepackage{xcolor}

\definecolor{codegreen}{rgb}{0,0.6,0}
\definecolor{codegray}{rgb}{0.5,0.5,0.5}
\definecolor{codepurple}{rgb}{0.58,0,0.82}
\definecolor{backcolour}{rgb}{0.95,0.95,0.92}

\lstdefinestyle{mystyle}{
    commentstyle=\color{codegreen},
    keywordstyle=\color{magenta},
    numberstyle=\tiny\color{codegray},
    stringstyle=\color{codepurple},
    basicstyle=\ttfamily\footnotesize,
    breakatwhitespace=false,         
    breaklines=false,                 
    captionpos=b,                    
    keepspaces=true,                 
    numbers=left,                    
    numbersep=5pt,                  
    showspaces=false,                
    showstringspaces=false,
    showtabs=false,                  
    tabsize=2
}

\title{Rethinking Soft Label in \\ Label Distribution Learning Perspective}

\author{%
  Seungbum Hong, Jihun Yoon, Bogyu Park, and Min-Kook Choi\thanks{Corresponding author.} \\
  AI Dev. Group\\
  Hutom\\
  Seoul, Republic of Korea \\
  \texttt{mkchoi@hutom.io} \\
}

\begin{document}

\maketitle

\begin{abstract}
The primary goal of training in early convolutional neural networks (CNN) is the higher generalization performance of the model. However, as the expected calibration error (ECE), which quantifies the explanatory power of model inference, was recently introduced, research on training models that can be explained is in progress. We hypothesized that a gap in supervision criteria during training and inference leads to overconfidence, and investigated that performing label distribution learning (LDL) would enhance the model calibration in CNN training. To verify this assumption, we used a simple LDL setting with recent data augmentation techniques. Based on a series of experiments, the following results are obtained: 1) State-of-the-art KD methods significantly impede model calibration. 2) Training using LDL with recent data augmentation can have excellent effects on model calibration and even in generalization performance. 3) Online LDL brings additional improvements in model calibration and accuracy with long training, especially in large-size models. Using the proposed approach, we simultaneously achieved a lower ECE and higher generalization performance for the image classification datasets CIFAR10, 100, STL10, and ImageNet. We performed several visualizations and analyses and witnessed several interesting behaviors in CNN training with the LDL.
\end{abstract}

\section{Introduction}

The supervision of a convolutional neural network (CNN) using a hard label has been very successful in most image classification problems [1, 2, 3]. However, in the training of a CNN using a hard label, as the number of weights of the network increases, [4] analyzed the overconfidence of the network prediction. To handle this phenomenon, [4] proposed the expectation of calibration error (ECE) to estimate the confidence of the model, and several approaches for calibrating the overconfidence of deep learning models were suggested, but they were not correlated with generalization performance and model calibration. Recently, several studies have introduced that data augmentation is effective for model generalization as well as calibration [5, 6], but the results are not significant in terms of generalization performance.

\begin{figure*}[t!]
\begin{center}
\includegraphics[width=0.9\linewidth]{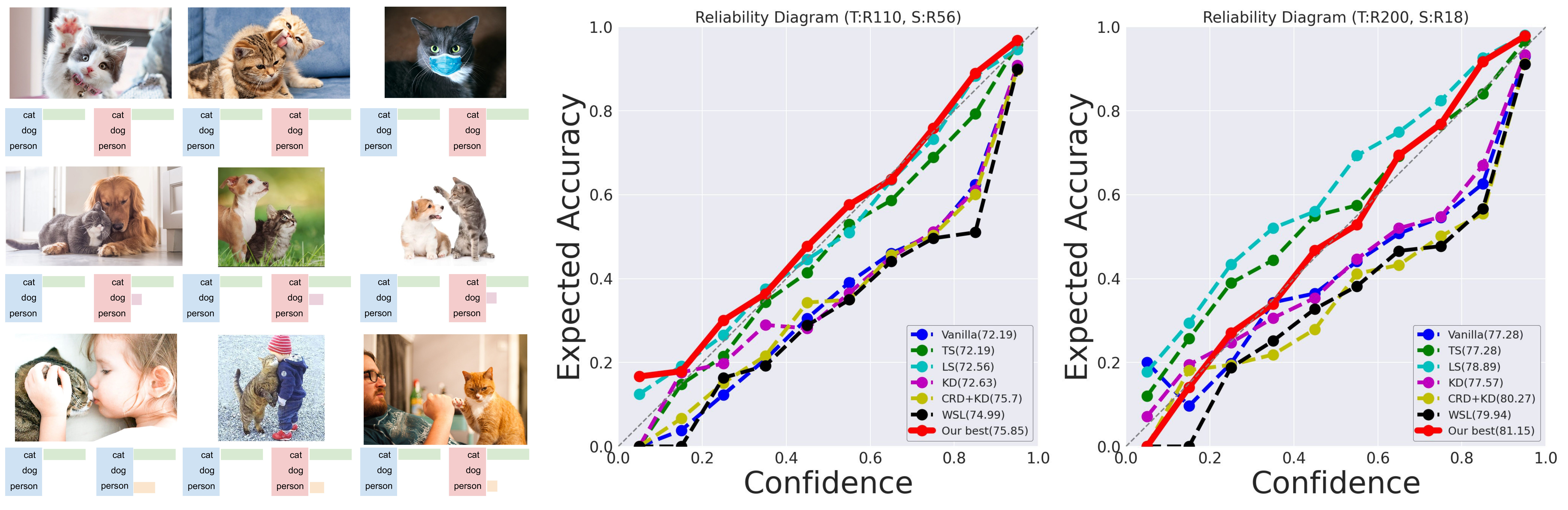}
\caption{\textbf{The difference between the supervision using the hard label (blue box) and using the soft label (red box) for image classification.} An illustration of the LDL for the cat class at the left is shown, and the reliability diagrams at the right show comparisons of the traditional hard label-based and LDL-based classification accuracy and ECE. Our LDL-based training successfully achieved better classification accuracy and lower ECE simultaneously.} 
\label{fig1}
\end{center}
\vspace{-7mm}
\end{figure*}

Label distribution learning (LDL) is designed for effective training through label distribution when the types of labels for supervision are difficult to define discretely and approaches the label generation (or enhancement) process as an optimization problem [20, 21]. Typically, LDL has been applied to applications that include inherent label ambiguity, such as facial age estimation, head pose estimation, facial emotion estimation, multi-label learning, partial multi-label learning, and video summarization [21, 22]. From a LDL perspective, label smoothing [14, 17, 19] is considered a subset of LDL.
We are inspired by the basic concepts of LDL and assume that the LDL potentially overcomes the discrepancy between the one-hot label-based training and the maximum confidence- based testing. To introduce this idea in a simple way, we exploited soft labels from teacher networks as a baseline distributed label. To learn the label distribution online differ from the former optimization approach, recent data augmentation techniques that merge labels and data during training were simultaneously applied. 


By applying the on/offline label distribution learning scenarios, we simultaneously obtained an improvement in the model generalization and calibration without additional regularization or architecture modification. The left section of Figure \ref{fig1} shows examples of the difference between hard label and LDL–based supervision for cat recognition. The graphs at the right of Figure \ref{fig1} show the reliability diagram [4] when different settings of modern KD approaches in the same CNN model are applied for CIFAR100. To verify the strength of LDL for model generalization and calibration, we performed a series of image classification tasks on datasets such as CIFAR10, 100 [23], STL10 [24], and ImageNet [25]. Based on a series of experiments with image classification, we confirmed that most recent KDs cause severe overconfidence, which impedes model calibration, and even a simple LDL approach can achieve better classification accuracy and suppression of model overconfidence.

\vspace{-2mm}
\section{Related Works}

\noindent \textbf{Model calibration.} ECE is an error that measures whether the prediction of the neural network can accurately estimate the true likelihood of the input data of the trained classes [4, 40]. In [4], temperature scaling was proposed in a way that can effectively be corrected and the reliability diagram is used to visualize model confidence for CNNs. In [30], various structural dropout methods and experiments on the drop rates according to each method were applied to the CNN model to analyze the correlation between model accuracy and ECE. VWCI [31] reduced the ECE and improved the recognition performance by defining a confidence integration loss as a probabilistic regularization term defined from a Bayesian model using multiple inferences based on probabilistic depth and dropout. It has also been reported that the AvUC loss based on uncertainty estimation in the model also aids in model calibration [32]. In addition to this, model training with mixup augmentation has been demonstrated to be effective in model calibration. However, it did not achieve much in improving the generalization performance of the model in preparation for the correction effect [5, 6]. \smallskip

\noindent \textbf{Label smoothing and label distribution learning.} Label smoothing was proposed to soften the hard label in the training process according to the given coefficient to prevent overconfidence and improve generalization performance [17, 18]. In [18], the authors analyzed the effect of label smoothing on deep neural network training by visualizing the penultimate fully connected layer of deep neural networks. According to the analysis results, there is evidence that the trained teacher network applied with label smoothing in the KD scenario can invalidate the effect of student model training. In recent studies [19, 28], the effect of label smoothing on teacher networks in the KD scenario was analyzed in more detail to extend the research results [18]. In [19], a quantification method that label smoothing erases meaningful information in the teacher network logit was proposed. In [28], the relationship between KD and label smoothing from the bias-variation perspective was analyzed. From the LDL perspective, Label smoothing can be regarded as a possible solution for LDL through constant softening of the hard label. We included label smoothing in our comparisons as one of the baselines to suppress overconfidence [18]. 

\noindent \textbf{Knowledge distillation.} Since the introduction of the KD [7], a vast number of approaches for knowledge distillation have been proposed [8-16, 26, 27]. FitNet [8] proposed a KD method that makes the feature maps of teacher networks similar. In recent years, various approaches have been proposed from the perspective of representation learning, such as RKD [10], which achieved transfer learning through geometric relations to the output of the model, CRD using metric learning [12], mutual learning based [13], self-supervised learning–based KD [15], and weighted soft label-based KD [16]. Among the variations of KD, the born-again network [9], which achieves transfer learning through repetitive training of student models without the use of a teacher network, and similar to [9], variations of KD that are free from teachers [14, 29] were also introduced. [18] and [19] explain the relationship between KD and label smoothing of teacher networks through empirical experiments. We argue that KD should be described in terms of LDL rather than label smoothing. We have observed that modern KDs spoil model calibration to improve generalization performance. 

\vspace{-3mm}

\section{On/Offline Label Distribution Learning (LDL)}

In this section, we briefly introduce the notations and approaches for KD and label smoothing (Section 3.1), which are basic prerequisites for our LDL-based approaches. Subsequently, on and offline approaches for LDL are described in Sections 3.2 and 3.3.

\subsection{Preliminaries}

\noindent \textbf{Knowledge distillation.} When the weight $w$ of the last fully connected layer for the $i$th feature input $x$ and the output with the softmax function for the $k$th class is given as $p_{i}^{k}=\frac{e^{(x_{i})^{T}w_k}}{\sum_{l=1}^{L}e^{(x_{i})^{T}w_l}}$, the softening output for the neural network is given by [7]. 

\setlength{\abovedisplayskip}{-3pt} \setlength{\abovedisplayshortskip}{-3pt}

\begin{align}
\bar{p}_{i}^k(x) = \frac{e^{((x_{i})^{T}w_k) \slash \tau}}{\sum_{l=1}^{L}e^{((x_{i})^{T}w_l) \slash \tau}},
\end{align}

\noindent where $\tau$ is a temperature scaling parameter that determines size of softening, and the total loss function $L=(1-\lambda)L_{CE}(y,p_{\theta_{s}})+\lambda L_{CE}(\bar{p}_{\theta_{t}},\bar{p}_{\theta_{s}})$ for the teacher model $\theta_t$ and the student model $\theta_s$, where $y$ is one-hot label and $L_{CE}(p,y)=-\sum_{k=1}^{K}y^k\log(p^k)$.

\noindent \textbf{Label smoothing} The label smoothing for the hard label $y_i$ for the same feature input is given as follows:

\begin{align}
\tilde{y}_{i}^{k}=(1-\alpha)y_{i}^{k}+\frac{\alpha}{K-1},
\end{align}

\noindent where $\alpha$ is the smoothing coefficient, and the probabilities for each class except the hard target corresponding to the $k$th class are evenly distributed as $\alpha/(K-1)$. \smallskip

\noindent \textbf{Expected calibration error.} The ECE for estimating the confidence of the neural networks proposed in [4] is estimated for $N_{test}$ samples when the softmax output $p_i$ inferred for all test data and the index with the maximum probability in the output is $\hat{c}_{i}=\underset{i}{\operatorname{argmax}}(p_i=k)$.

\begin{align}
ECE= \sum_{m=1}^{M}\frac{|H_m|}{N_{test}}\left( \frac{1}{|H_m|} \sum_{i \in H_m} 1(\hat{c}_{i}=c_i) - p_i\right),
\end{align}

\noindent where $H_m$ is the index set and generates $M$ interval bins of $((m-1)/M, m/M ]$ for $N_{test}$ samples. Typically ECE is measured by the histogram for a bin of 0.1 size by setting $M=10$. \smallskip

\begin{figure*}[t!]
\begin{center}
\includegraphics[width=1.0\linewidth]{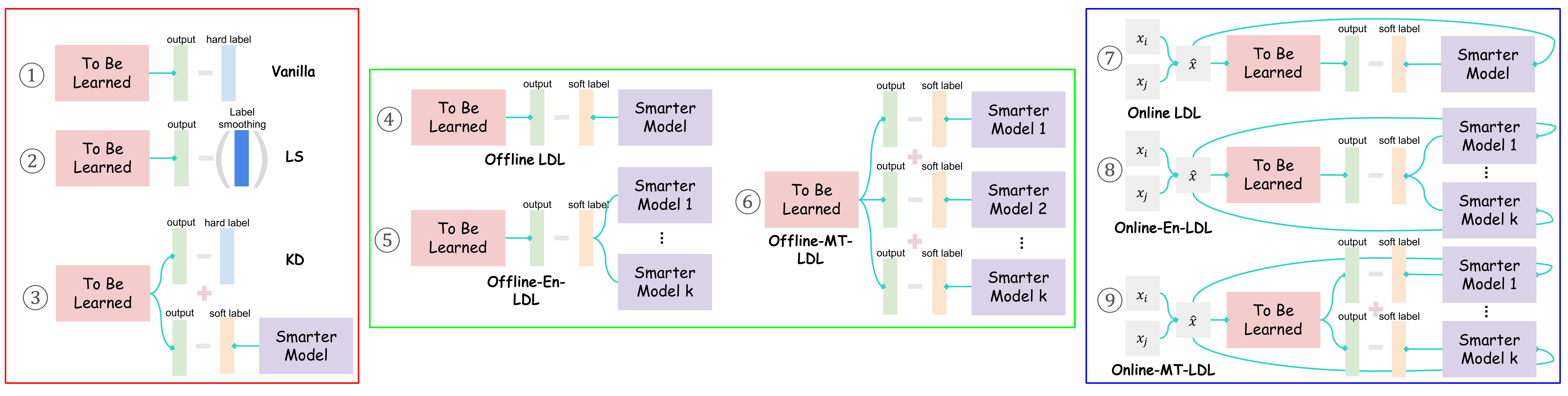}
\caption{\textbf{Schematic of the training configurations.} We used abbreviations to simplify the notation: \#1 learning from scratch (Vanilla), \#2 label smoothing (LS), \#3 knowledge distillation (KD), \#4 soft label (Off-LDL), \#5 teacher ensemble for soft label (Off-En-LDL), \#6 linear combination with the multiple soft label (Off-MT-LLD), \#7 soft label with data augmentation (On-LDL), \#8 soft label using data augmentation with teacher ensemble (On-En-LDL), \#9 linear combination of multiple soft label using data augmentation (On-MT-LDL). The red box represents the existing training method, the green box represents the offline approaches, and the blue box represents the online approaches.} 
\label{fig2}
\end{center}
\vspace{-11mm}
\end{figure*}

\noindent \textbf{Criterion of LDL with cross entropy loss.} The main objective of cross entropy loss with LDL perspective is given by:

\begin{align}
L_{CE}(p,z)=-\sum_{k=1}^{K}z^k\log(p^k),
\end{align}

\noindent where $z$ is a label vector that satisfies $\sum_j z_j=1$. Label smoothing or soft label by output of teacher networks can be regarded as a specific solution for $z$ ($z=\tilde{y}(\alpha)$ or $z=\bar{p}_{\theta_{t}}^{k}$). We reformulated the problem $\text{argmax}_\theta (E[h_{D_{train}}(x,y;\theta)]-E[h_{D_{test}}(x,y;\theta)])$ to find the CNN with the maximum generalization performance in a specific image classification dataset $D \ni \{D_{train},D_{test}\} $ as follows:

\begin{align}
\operatorname*{argmax}_{\theta, Z} E[h_{D_{train}}(x,z;\theta)]-E[h_{D_{test}}(x,y;\theta)],
\end{align}

\noindent where $Z_{D_{train}} \ni \{ z_1, ..., z_{N_{train}} \}$ is a set of new labels for all training data. Equation (5) can be solved as an optimization problem of finding a pair of the optimal labels for each input data $(x_i,z_i)$ such as the previously proposed LDL approaches [20, 21]. Since traditional approaches cannot update $z$ and $\theta$ simultaneously during deep neural network training, we applied simple but effective on/offline approaches. For the simplicity, we used basic KD setting, which teacher network generate new label set $Z$ for target (student) neural network training. The Offline setting is a way to generate $Z$ as a soft label as the output of the teacher network. $Z$ is not updated during training, but some variations are possible depending on the way the ensemble of teacher networks. The online setting is a way of continuously transforming $Z$ while updating $\theta$. Since it is difficult to presume an optimal $Z$, we generated diverse labels with modern data augmentation techniques. As with offline settings, there are several variations depending on the way the ensemble of teacher networks and label generation process. Figure \ref{fig2} shows different types of training configurations for baseline and LDL. 

\subsection{Offline LDL}

We simplified the problem by using the KD setting by the teacher output as the new label to extract feasible solutions for each sample pair $(x_i,z_i;\theta)$. The offline LDL is illustrated in the green box in Figure \ref{fig2}, such that the set of sample pairs $(X_{D_{train}}, Z_{D_{train}})$ under $X_{D_{train}} \ni \{x_1, ..., x_{N_{train}}\}$ is fixed during the training process. The cross-entropy loss for training with the label generated by the teacher $\theta_t$ and the student model to be trained is $\theta_s$ is as follows:

\begin{align}
L_{CE}(p,\bar{z})=-\sum_{k=1}^{K}\bar{z}_i^{k}\log(p_{i,\theta_{s}}^{k}),
\end{align}

\noindent where $\bar{z}$ is defined by way of generating new labels. We used simple variations of $\bar{z}=f(\cdot)$ for offline LDL (see Figure \ref{fig2}): soft label $f(x_i^k;\theta_{t})$ (Off-LDL, \#4), soft label with teacher ensemble $\bar{z}=\frac{1}{N}\sum_{n=1}^N{f(x_i^k;\theta_{t,n})}$, where $N$ is the number of teacher models (Off-En-LDL, \#5), and linear combination of soft labels from multiple teachers $-\sum_{n=1}^N\sum_{k=1}^{K}\bar{z}_{i,\theta_{t,n}}^{k}\log(p_{i,\theta_{s}}^{k})$ (Off-MT-LDL, \#6).

\subsection{Online LDL}

To reflect the objective of Equation (5) during training, it is necessary to update $Z$ and $\theta$ simultaneously. We applied recent data augmentation techniques that can simply adopt online LDL. Among the proposed data augmentation techniques, some techniques manipulate input data and label together. A generalized form of augmentation technique considering data and labels together is

\begin{gather}
\hat{x}= \mathrm{M}_1\otimes x_1 + ... + \mathrm{M}_P \otimes x_P \nonumber \\
\hat{y}= \lambda_1y_1 + ... + \lambda_Py_P,
\end{gather}

\noindent where $\hat{x}$ is an augmented sample of mixed label $\hat{y}$ with $\sum^{P}{\lambda_i = 1}$ up to $P$ samples. $\mathrm{M}$ is a blending mask equal to the data width and height, and satisfies $\sum^{P}{\mathrm{M}_i(u,v)=1}$, where $u$ and $v$ indicate the pixel location. Each augmentation algorithm is designed to stochastically determine the location and size of each sample for blending and mainly follows a uniform distribution. $\mathrm{M}$ is provided differently for each augmentation technique. Typically, When $P=2$, $x_1$ is defined as the target image, and $x_2$ is defined as the 0 image for CutOut [3].  We exploited mixup [33], CutMix [34], and RICAP [35] for data augmentation, and online LDL with data augmentation was as follows:

\begin{align}
L_{CE}(\hat{p},\hat{z})=-\sum_{k=1}^{K}\hat{z}_i^{k}\log(\hat{p}_{i,\theta_{s}}^{k}),
\end{align}

\noindent where $\hat{p}_{i}^{k}=\frac{e^{(\hat{x}_{i})^{T}w_k}}{\sum_{l=1}^{L}e^{(\hat{x}_{i})^{T}w_l}}$. Data augmentation applies equally to teacher models for label enhancement $\hat{z}_i^k$, but the mixed label $\hat{y}$ is not used for training. Similar to the offline approaches corresponding to \#5 and \#6 in Figure \ref{fig2}, online LDL can be easily extended. The enhanced label through the augmentation-based teacher ensemble is obtained as $\hat{z}=\frac{1}{N}\sum_{n=1}^N{f(\hat{x}_i^k;\theta_{t,n})}$ (On-En-LDL, \#8), and the linear combination of the augmentation-based soft labels from multiple teachers is given as $-\sum_{n=1}^N\sum_{k=1}^{K}\hat{z}_{i,\theta_{t,n}}^{k}\log(\hat{p}_{i,\theta_{s}}^{k})$ (On-MT-LDL, \#9).

\section{Experimental Results}

\noindent \textbf{Experimental setting.} We performed a series of experiments on the CIFAR10, 100 [23], and STL10 [24] datasets to verify the performance of the on/offline LDL. First, the major experiments were performed with the teacher-student network configurations of the well-used ResNet architectures [3]. We divided ResNet into small and large networks according to model size. Small size models include ResNet20 (0.27M), 56 (0.85M), and 110 (1.7M) and large size models include ResNet18 (11.18M), 50 (23.51M), and 200 (62.62M). For CIFAR10, 100, and STL10, a total of 240 epochs was trained to start with an initial learning rate of 0.05, and 0.1 learning rate scaling was applied at 150, 180, and 210 epochs. The weight decay was set to $5.0 \times 10^{-4}$, the batch size was set to 64. We also tested a long training scenario with a basic training configuration, with the assumption that LDL can fundamentally achieve better performance when the number of training pairs of sample and label is large especially in online LDL. For long training, a total of 350 epochs was trained to start with an initial learning rate of 0.05, and 0.1 learning rate scaling was applied at 150, 200, 250, and 300 epochs\footnote{The long training is marked with  '+'.}. In all ensemble (En) and multiple teacher (MT) settings, ResNet20, 32, 44, 56, and 110 used together, were trained by the same learning scheduler. We measured the image classification accuracy and ECE [4] to evaluate the performance. Visualization of reliability diagrams is provided to intuitively check the strength of model calibration of the network in the same way as in [4]. \smallskip

\noindent \textbf{LDL with data augmentation.} The augmentation algorithms applied for On-LDL are mixup [33], CutMix [34], and RICAP [35], and the default hyper-parameter for data augmentation refers to the original implementation of each algorithm. Table \ref{data_augmentation} shows the recognition results of the on/offline LDL according to each data augmentation technique. All three algorithms showed rather poor performance in small networks and were able to achieve significant performance improvement in large networks. For long training, we only report the LDL of the augmentation technique with the best performance. All training was performed on 3 different random seeds. We omit variations in calibration scores, which are no significant differences. We selected the CutMix [34] or RICAP [35] as a data augmentation for remain LDL experiments. Similar to the results of [5], with the mixup augmentation, the improvement in accuracy was not significant, but a model calibration effect could be seen in some cases. Rather, CutMix and RICAP achieved steady performance improvement and better model calibration. Offline LDL achieved the best accuracy and model calibration performance in ResNet20, and online LDL improved significantly as the size of the network increased. With the ResNet50, an accuracy improvement of up to 1.9\% and better model calibration was achieved than in the case of data augmentation only. \smallskip

\begin{table}[t!]
\caption{Classification accuracy (\%) and ECE of vanilla and each LDL setup for CIFAR100 depending on each data augmentation methods [33, 34, 35]. }
\begin{center}
\label{data_augmentation}
\scalebox{0.75}{
\begin{tabular}{ccccc}

\hline
Teacher 			& ResNet110 					& ResNet110 					& ResNet200					& ResNet200			\\ 
Student (\# param)	& ResNet20 (0.27M)				& ResNet56 (0.85M)				& ResNet18 (11.18M)			& ResNet50 (23.51M)	\\ \hline
Vanilla 			& 69.32±0.27/0.070				& 72.28±0.09/0.123				& 77.83±0.55/0.080				& 78.80±0.17/0.107 	\\ 
Vanilla [33] 		& 67.29±0.17/0.127				& 73.10±0.11/0.118				& 78.71±0.29/0.131				& 79.37±0.51/0.059	 	\\ 
Vanilla [34]		& 67.23±0.28/0.075				& 73.99±0.16/0.066				& 80.32±0.18/0.045				& 81.73±0.07/0.038 	\\ 
Vanilla [35]		& 68.46±0.12/0.070				& 73.95±0.17/0.027				& 80.10±0.05/0.039				& 81.47±0.34/0.039	 	\\ \hline \hline
Off-LDL			& \textbf{69.9±0.19}/\textbf{0.051}	& 73.59±0.36/0.085				& 78.67±0.14/0.060				& 79.19±0.37/0.087		\\ 
On-LDL [33]		& 68.42±0.12/0.130				& 73.73±0.67/0.115				& 79.76±0.31/0.121				& 81.03±0.01/0.051		\\
On-LDL [34]		& 68.25±0.06/0.073				& 74.44±0.16/0.060				& 81.26±0.11/0.043				& 83.09±0.05/\textbf{0.030}	\\
On-LDL [35]		& 68.90±0.05/0.063				& 74.87±0.13/0.025				& 80.57±0.04/0.056				& 81.64±0.06/0.039		\\ 
On-LDL+ 			& 69.41±0.17/0.073				& \textbf{75.56±0.28}/\textbf{0.022} 	& \textbf{81.76±0.25}/\textbf{0.034}	& \textbf{83.57±0.05}/0.038	\\ \hline
\end{tabular}
}
\end{center}
\vspace{-2mm}
\end{table}

\noindent \textbf{CIFAR10 and STL10.} Table \ref{small_cifar10_stl10} shows the evaluation results of the CIFAR10 and STL10 datasets for LDL training. In the two relatively small datasets, we did not evaluate the large network, as only the small network could provide sufficient performance improvement. In both CIFAR10 and SLT10, online LDL utilizing multiple teacher networks showed improvements in accuracy and model calibration. In CIFAR10, the accuracy increase of up to 1.6\% and the ECE reduction effect of up to about 80\% were obtained in ResNet56 compared to the vanilla model. In STL10, the accuracy increase of 3.39\% and ECE improvement effect of about 85\% were obtained in ResNet20 compared to the vanilla model. In CIFAR10 and STL10, RICAP achieved better performance than CutMix.

\begin{table}[t!]
\caption{Classification accuracy and ECE for small size ResNets for CIFAR10 and STL10.}
\begin{center}
\label{small_cifar10_stl10}
\scalebox{0.75}{
\begin{tabular}{c|cc|cc}
						& CIFAR10					& 						& STL10						&						\\ \hline
	 					& Model						& 						& Model						&						\\
Method					& ResNet20					& ResNet56				& ResNet20					& ResNet56				\\ \hline
Vanilla	 				& 92.56±0.10/0.033				& 93.88±0.08/0.038 	 		& 83.44±0.10/0.067				& 84.15±0.31/0.074			\\ 
Label smoothing			& 92.41±0.25/0.052				& 93.72±0.19/0.063			& 83.54±0.22/0.111				& 84.35±0.01/0.091			\\ 
KD ($\alpha$=0.1,$T$=3) [7]	& 92.56±0.02/0.032				& 93.94±0.03/0.038			& 83.95±0.56/0.070				& 84.62±0.41/0.070			\\ \hline
Off-LDL					& 92.62±0.18/0.028				& 93.96±0.16/0.031			& 84.14±0.42/0.055				& 84.57±0.13/0.054			\\ 
Off-En-LDL			 	& 92.60±0.17/0.031				& 94.07±0.01/0.032 			& 83.40±0.01/0.067				& 84.03±1.07/0.069			\\ 
Off-MT-LDL			 	& 93.05±0.28/0.022				& 94.15±0.17/0.022			& 84.16±0.23/0.054				& 84.52±0.09/0.060			\\ \hline
On-LDL		 			& 92.92±0.08/0.009				& 94.04±0.09/0.013			& 85.97±0.12/0.023				& 86.24±0.37/0.029			\\
On-LDL+					& 93.33±0.02/0.008				& 94.32±0.05/0.011			& 85.98±0.04/0.023				& 86.61±0.15/0.029			\\ 
On-En-LDL		 		& 93.25±0.06/0.016				& \textbf{94.48±0.09}/0.016	& 86.45±0.17/0.030				& 86.92±0.42/0.034			\\
On-MT-LDL				& 93.14±0.05/0.006				& 94.44±0.07/\textbf{0.007}	& 86.44±0.14/\textbf{0.008}		& 87.08±0.06/\textbf{0.009}	\\ 
On-En-LDL+				& 93.71±0.02/0.017				& 94.46±0.27/0.017			& 86.82±0.03/0.030				& \textbf{87.13±0.08}/0.037	\\
On-MT-LDL+				& \textbf{94.03±0.08}/\textbf{0.006}	& 94.41±0.02/0.006			& \textbf{86.93±0.02}/\textbf{0.008}	& 86.92±0.08/0.010			\\ \hline
\end{tabular}
}
\end{center}
\vspace{-4mm}
\end{table}

\begin{figure}[t!]
\begin{center}
\includegraphics[width=0.7\linewidth]{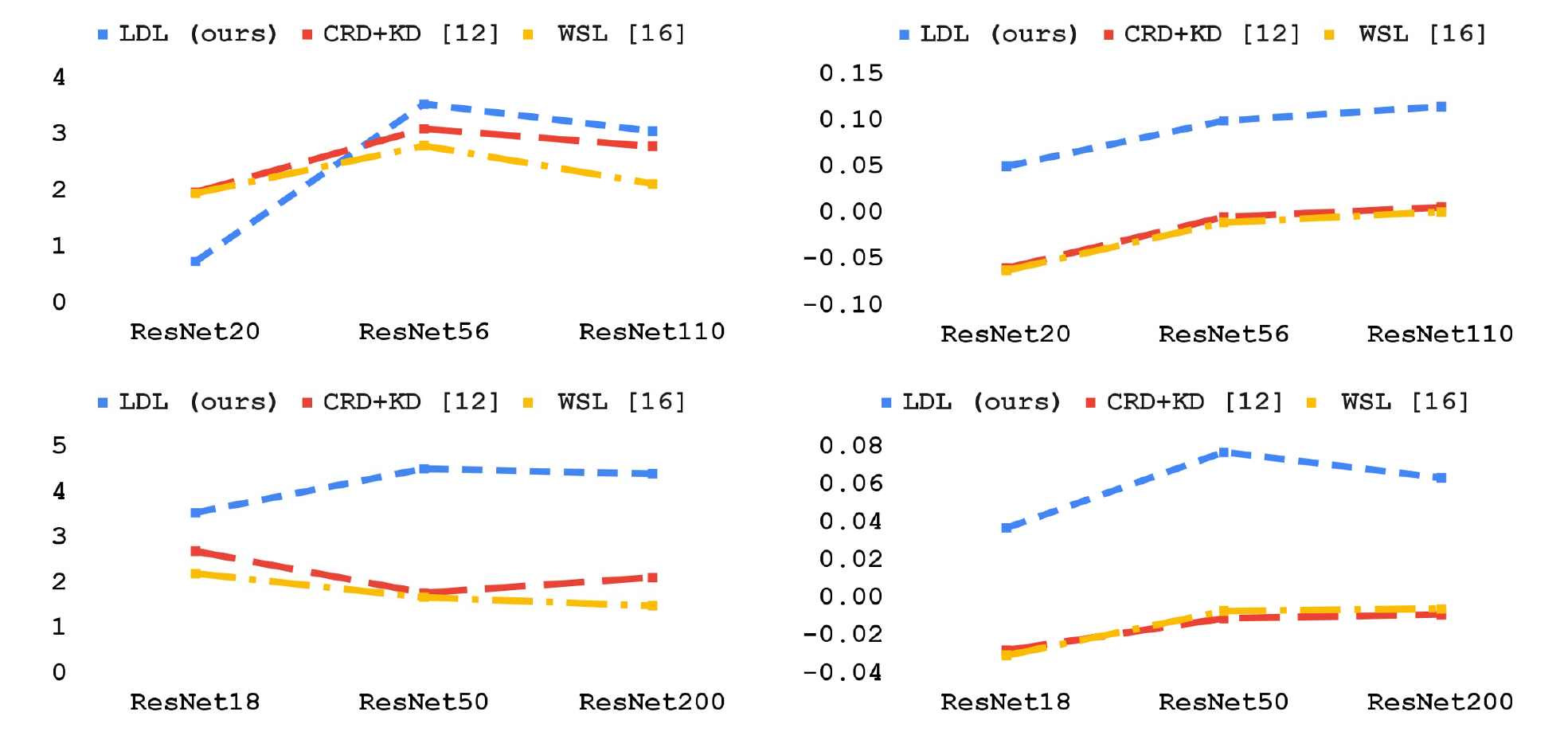}
\caption{\textbf{Performance change by methods according to small and large student models for CIFAR100.} The left graph shows the accuracy difference with the vanilla training by the number of weights of the student model, and the right graph shows the ECE difference with the vanilla training. The LDL technique shows continuous improvement in model correction as the accuracy increases, and the effect increases as the model size increases. On the other hand, SOTA KD methods have a steady improvement in accuracy, but rather impede model reliability.} 
\label{graph}
\end{center}
\vspace{-5mm}
\end{figure}

\noindent \textbf{CIFAR100.} Table \ref{cifar100} shows the evaluation results of LDL in CIFAR100. In the upper part of Table \ref{cifar100}, the improvement in generalization performance is relatively insignificant for small networks. We hypothesized that a large number of weights is required to sufficiently train the LDL-based label variants and evaluated the small and large networks simultaneously on the CIFAR100 dataset. When the model has a large number of weights, it is well trained on the label variation of the new label distribution, and it is possible to achieve sufficient performance improvement and model calibration without the multiple teachers. We obtained two main observations from the experiments with three benchmarks: 1) The classification accuracy is mainly determined by the number of weights with LDL approaches, and the influence of the number of parameters in the teacher model is not noticeable. 2) The effectiveness of model calibration steadily improves regardless of the number of parameters in each model.
We also compared the proposed LDL approaches with the SOTA KD methods [12, 15, 16] on the CIFAR100 dataset. Table \ref{cifar100} shows the comparison results with small and large student networks in terms of classification accuracy and ECE. In most cases, LDL-based methods achieved a higher generalization performance and lower ECE simultaneously. Figure \ref{graph} shows this phenomenon more dramatically: as the number of parameters in the student network is small, the SOTA KD methods show a very high ECE score, have no explanatory power for model confidence, and result in overconfidence. LDL techniques were able to achieve better model calibration compared to SOTA KDs as the number of parameters in the student network increased. Classification accuracy showed up to 2.86\% improvement in ResNet200 compared to the best performing CRD+KD, and an improvement of at least 60\% up to 88\% compared to the SOTA KD methods in model calibration\footnote{Types of teacher networks, long training results of SOTA KD methods, loss curves, and additional experimental results and analysis are in the supplementary material.}. \smallskip

\begin{table}[t!]
\caption{Classification accuracy and ECE of the LDL and the SOTA KDs for the CIFAR 100. Among the LDL techniques, the performance of the models that achieved the highest accuracy or the lowest ECE were recorded.}
\begin{center}
\label{cifar100}
\scalebox{0.75}{
\begin{tabular}{cccc}
\hline
Teacher 					& 							& ResNet110					& 				\\ 
Student					& ResNet20					& ResNet56					& ResNet110		\\ \hline
Vanilla 					& 69.32±0.27/0.070				& 72.28±0.09/0.123				& 73.88±0.15/0.131 \\ 
Label smoothing			& 69.43±0.20/0.053				& 72.87±0.17/\textbf{0.020}		& 73.90±0.16/0.051 \\ \hline
KD ($\alpha$=0.1,$T$=3) [7]	& 69.07±0.23/0.071				& 72.76±0.11/0.118				& 73.60±0.16/0.135 \\ 
CRD	[12]					& 71.05±0.12/0.060				& 74.82±0.11/0.107				& 76.04±0.04/0.121 \\ 
CRD+KD [12]				& \textbf{71.29±0.23}/0.129		& 75.38±0.27/0.127				& 76.67±0.46/0.125 \\ 
SSKD [15]					& 71.00±0.04/0.122				& 74.88±0.14/0.119				& 75.73±0.17/0.118 \\ 
WSL	[16]	 				& 71.27±0.26/0.132				& 75.08±0.19/0.133				& 76.00±0.52/0.130 \\ \hline
Off-MT-LDL				& 70.75±0.15/\textbf{0.019}		& 74.84±0.04/0.023				& 76.35±0.18/\textbf{0.016} \\ 
On-LDL+					& 69.41±0.17/0.073				& \textbf{75.54±0.24}/0.029		& \textbf{77.28±0.20}/0.025 \\ \hline	\hline	
Teacher 					& 							& ResNet200					& 				\\ 
Student					& ResNet18					& ResNet50					& ResNet200		\\ \hline
Vanilla 					& 77.83±0.55/0.080				& 78.57±0.35/0.107				& 79.47±0.58/0.101 \\ 
Label smoothing			& 78.95±0.06/0.085				& 78.90±0.24/0.042				& 79.59±0.42/\textbf{0.037} \\ \hline
KD ($\alpha$=0.1,$T$=3) [7]	& 77.73±0.16/0.077				& 79.12±0.45/0.103				& 80.10±0.23/0.100 \\ 
CRD+KD [12]				& 80.41±0.14/0.108				& 80.34±0.07/0.118				& 81.58±0.20/0.110 \\ 
WSL	[16]			 		& 79.91±0.03/0.111 				& 80.25±0.12/0.114				& 76.00±0.52/0.130 \\ \hline
On-LDL					& 81.26±0.11/0.043				& 83.09±0.05/\textbf{0.030}		& 83.88±0.28/0.038 \\ 
On-LDL+					& \textbf{81.76±0.25}/\textbf{0.034}	& \textbf{83.57±0.05}/0.038		& \textbf{84.44±0.13}/0.039 \\ \hline	
\end{tabular}
}
\end{center}
\vspace{-3mm}
\end{table}
 
\noindent \textbf{ImageNet.} We evaluated the ImageNet dataset [25] to verify LDL methods on the large-scale dataset. For the evaluation of the ImageNet, we set up the multiple teacher and student network configurations. The learning scheduler applied the configuration of \cite{He16}, and the on/offline LDL methods were tested. As shown in Table \ref{imagenet}, similar to the other datasets, the similar tendency of improved accuracy and model calibration was achieved simultaneously compared to vanilla and label smoothing.\smallskip

\begin{table}[t!]
\caption{Classification accuracy (top1 / top5) and ECE for ImageNet dataset depending on each label enhancement setup.}
\begin{center}
\label{imagenet}
\scalebox{0.75}{
\begin{tabular}{ccccc}
\hline
Teacher 			& ResNet152							& ResNet152				& ResNet152					& ResNet152					\\ 
Student			& ResNet18 							& ResNet50				& ResNet101					& ResNet152 					\\ \hline
Vanilla			& 70.39/89.54, 0.014						& 75.96/92.81, 0.032			& 77.43/93.72, 0.045				& 78.28/94.14, 0.050 			\\ 
Label smoothing	& 70.40/89.52, 0.102						& 76.56/93.12, 0.070			& 78.36/94.05, 0.058				& 78.82/94.30, 0.036				\\ \hline
Off-LDL			& \textbf{71.63}/\textbf{90.46}, \textbf{0.013}	& 77.27/\textbf{93.61}, 0.021	& 78.72/\textbf{94.30}, 0.024		& 79.16/94.55, \textbf{0.021}		\\ 
On-LDL			& 69.03/88.91, 0.078						& 75.84/92.99, \textbf{0.017}	& \textbf{78.97}/94.27, \textbf{0.016} 	& \textbf{79.51}/\textbf{94.72}, 0.022 \\ 
On-LDL+			& 69.43/89.20, 0.062						& \textbf{77.30}/93.54, 0.023	& 78.82/94.28, 0.022				& 79.25/94.49, \textbf{0.021}		\\ \hline
\end{tabular}
}
\end{center}
\vspace{-4mm}
\end{table}

\noindent \textbf{Other model architectures.} To validate the effect of LDL in various architectures, we performed an evaluation according to ResNet200 teacher and ResNeXt [36], DenseNet [37], and DLA [38] student networks in the CIFAR100. Table \ref{other_architectures} lists the accuracy and model calibration improvements by LDL approaches. Compared to the vanilla model in other types of architectures, there was an accuracy improvement of about 3-5\% and an ECE reduction of 20-66\% was achieved.
\smallskip

\begin{table*}[t!]
\caption{Evaluation of different types of architectures.}
\begin{center}
\label{other_architectures}
\scalebox{0.8}{
\begin{tabular}{cccc}
\hline
Method 					& 							& Student (\# param)				&		 	 				\\ 
 						& ResNext29\_4x64d (27.1M) [36] 	& DenseNet121 (6.95M) [37]		& DLA (16.29M) [38] 				\\ \hline
Vanilla	 				& 80.30±0.14/0.050				& 79.67±0.26/0.081				& 77.27±0.50/0.099 				\\ 
Label smoothing			& 80.73±0.37/0.151				& 79.81±0.28/0.044				& 79.07±0.33/0.039 				\\ 
KD ($\alpha$=0.1,$T$=3) [7]	& 80.38±0.34/0.052				& 79.53±0.10/0.080				& 77.67±0.21/0.101		 		\\ \hline
Off-LDL	 				& 80.59±0.10/\textbf{0.039}		& 80.29±0.20/0.062				& 78.12±0.06/0.081	 			\\ 
On-LDL			 		& \textbf{84.16±0.13}/0.059		& \textbf{83.35±0.21}/\textbf{0.028}	& \textbf{82.93±0.08}/\textbf{0.033}	\\ \hline
\end{tabular}
}
\end{center}
\vspace{-5mm}
\end{table*}

\begin{figure*}[t!]
\begin{center}
\includegraphics[width=0.9\linewidth]{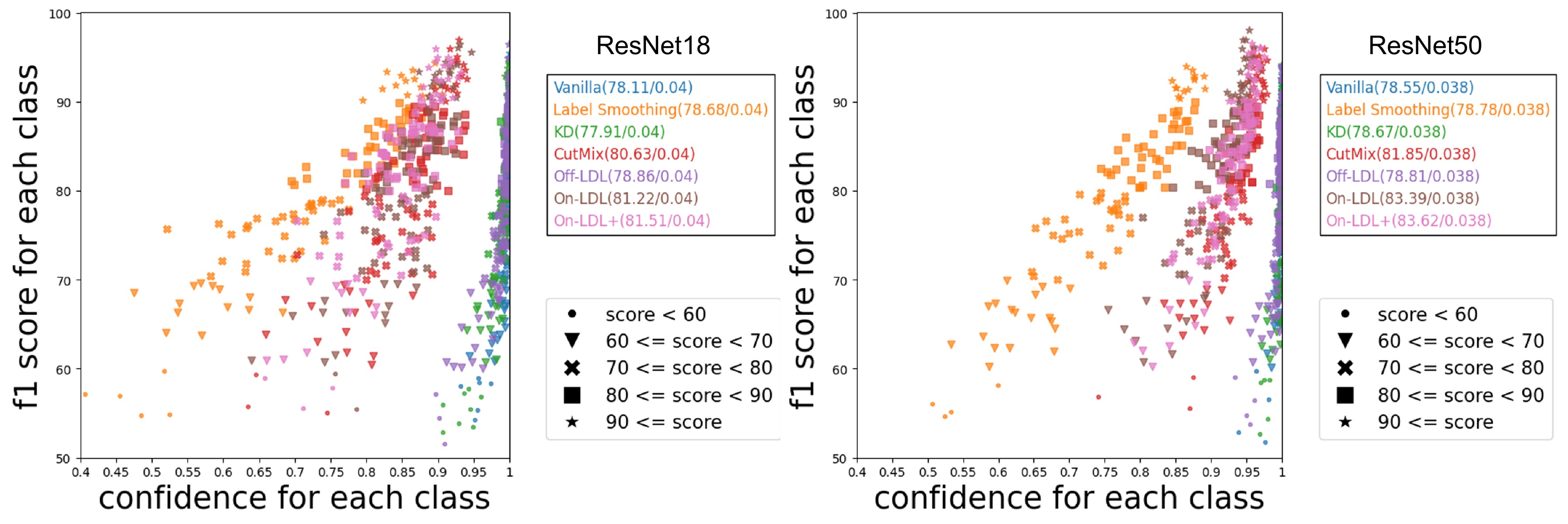}
\caption{\textbf{Plotting the relationship between the F1 score and the average output confidence for the GT class of the model according to the training methodology.} The average output confidence for each of the 100 classes of CIFAR100 and different shapes are marked for each F1 score. Values in parentheses are the accuracy and ECE of each method.} 
\label{f1_confidence}
\end{center}
\vspace{-3mm}
\end{figure*}

\begin{figure*}[t!]
\begin{center}
\vspace{-1mm}
\includegraphics[width=1.0\linewidth]{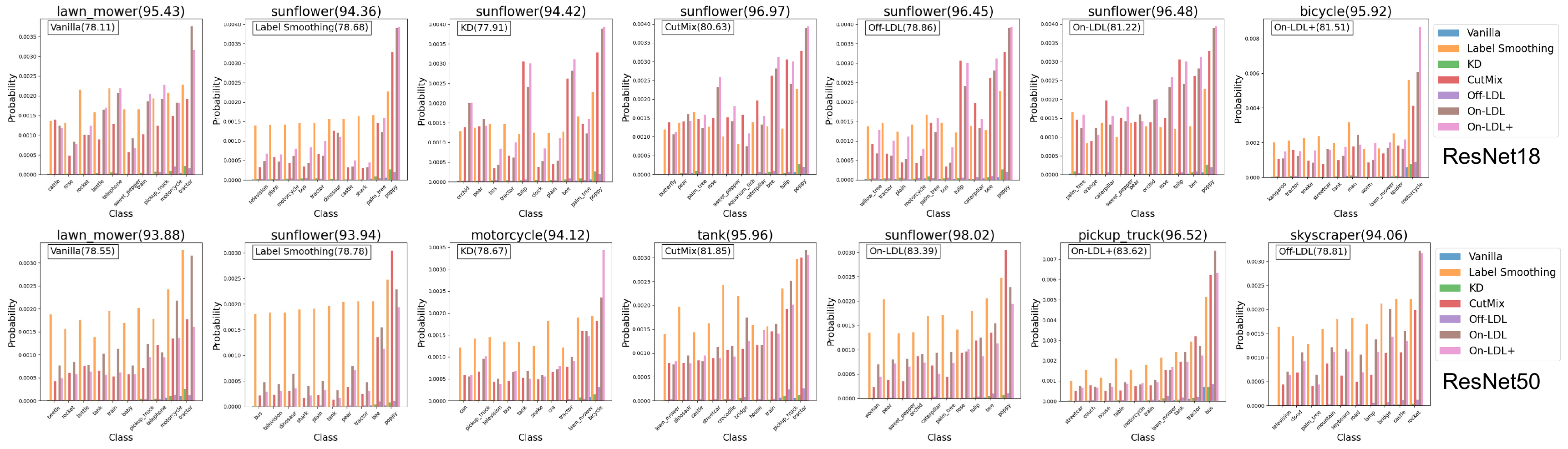}
\caption{\textbf{The top 10 high-probability classes according to the best-performing classes by training methodology.} For each method, the F1 score for the best performing class was recorded next to the class name and overall accuracy was recorded next to the method name.} 
\label{class_plot}
\end{center}
\vspace{-3mm}
\end{figure*}

\begin{figure*}[t!]
\begin{center}
\vspace{-1mm}
\includegraphics[width=0.65\linewidth]{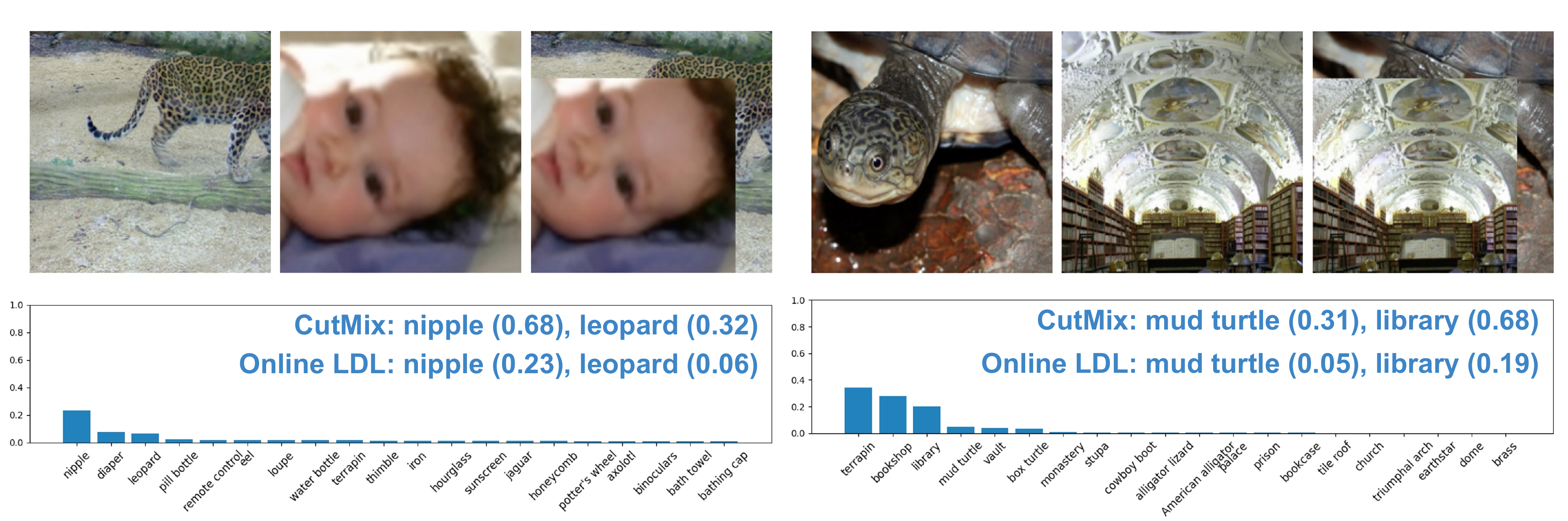}
\caption{\textbf{Examples of supervision by CutMix and LDL.} These examples from ImageNet are label distribution and input sample pairs obtained from a teacher network after CutMix augmentation.}  
\label{ldl_example} 
\end{center}
\vspace{-3mm}
\end{figure*}

\noindent \textbf{Why does LDL work?} Based on the evaluation results of four datasets, LDL is observed to have an excellent effect on generalization performance and model calibration for CNN training. Figure \ref{f1_confidence} is the result of the test set plotting the average confidence of ResNet18 and 50 for the ground truth class on the x-axis and the F1 score for each class on the y-axis. In the case of one-hot label-based training or offline LDL using only soft labels of a teacher trained on the one-hot label, most have an average confidence score of 0.9 or higher regardless of the F1 score. Applying data augmentation or label smoothing alleviates this over-confidence, but in the case of label smoothing, the confidence distribution has an excessively large variance as a result of the forced softening. The case of online LDL appears to achieve effective model calibration by balancing the distribution of confidence on the output. Figure \ref{class_plot} plots the top 10 highest-probability classes for the best-performing class for each training method for CIFAR100. Not only does the training methodology change the best performing class, but the distribution of output confidence for each class is also very different. The results of over-softening of label smoothing and over-confidence in the one-hot label are also observed here. Figure \ref{ldl_example} shows examples of soft labels obtained from the teacher network after the input sample undergoes data augmentation in training on the ImageNet. We observed that the distribution of labels supervising student networks differed significantly from that of the CutMix.

\begin{figure*}[t!]
\begin{center}
\vspace{-1mm}
\includegraphics[width=0.9\linewidth]{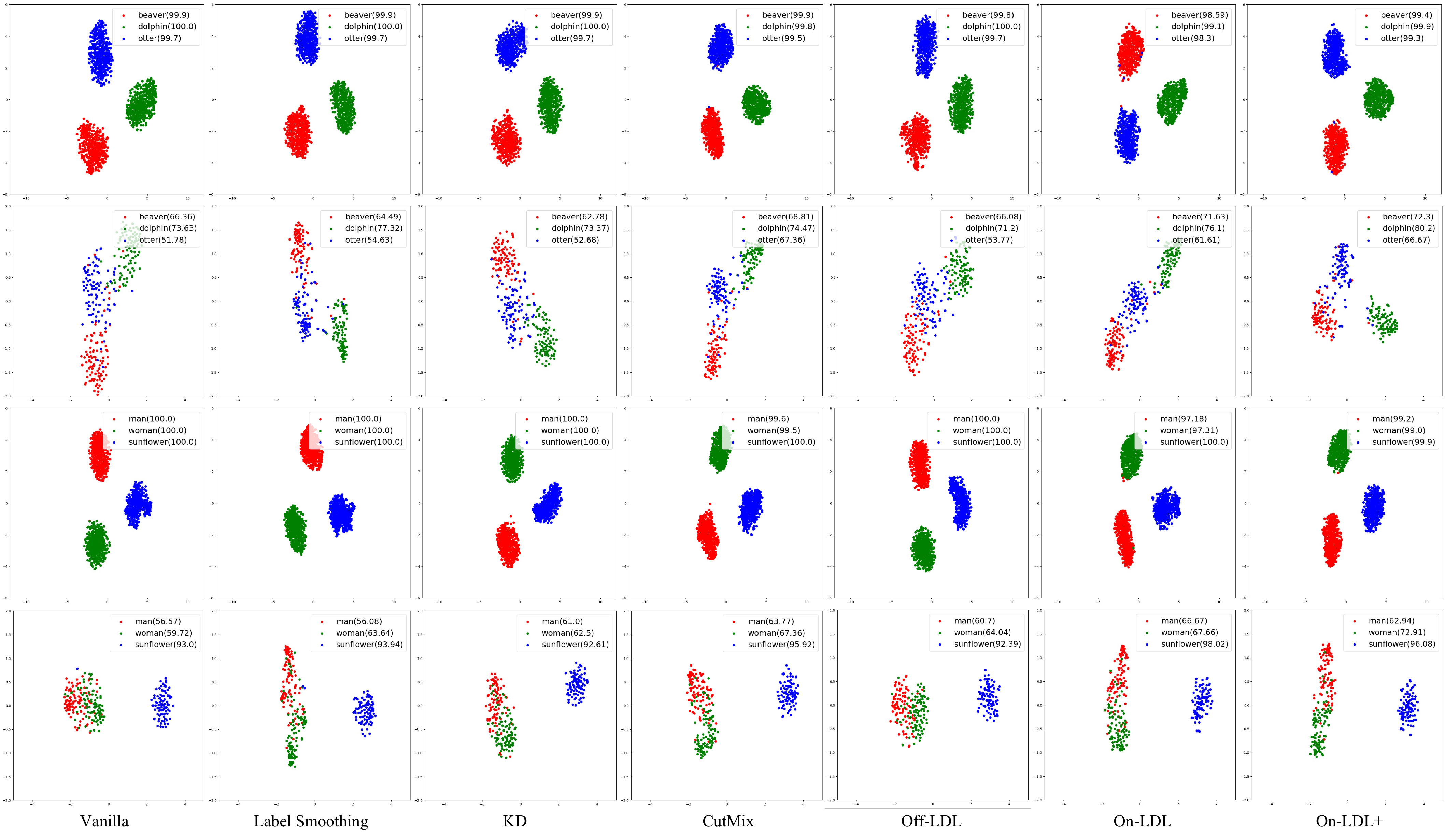}
\caption{\textbf{Visualization of penultimate layer's activation.} The first row is the training samples of the 'beaver', 'otter', and 'dolphin' classes, and the second row is the test samples. The third row is the training samples of the 'man', 'woman', and 'sunflower' classes, and the last row is the test samples. All plots are visualization of the activation of ResNet50.}  
\label{penultimate_output} 
\end{center}
\vspace{-7mm}
\end{figure*}

\noindent \textbf{Penultimate layer output visualization.} We plot the activation of the penultimate layer to visualize the effect of LDL on the feature representation as in [18, 19]. 'beaver' and 'otter' are semantically similar classes and 'dolphin' are semantically different classes. Very interestingly, it is observed that LDL plays an appropriate role in the classification of semantically similar classes. Looking at the first row of Figure \ref{penultimate_output}, the online LDL has a geometry of activation that is more effective for semantically similar classification when long training is performed. Similarly, for the relationships of 'man', 'woman', and 'sunflower', although less prominent than in the previous example, online LDL produces more efficient geometries for classification than other methods.

\vspace{-3mm}
\section{Conclusion}
\vspace{-2mm}
We observed and analyzed the effects of label smoothing, KD, and data augmentation on classification accuracy and model confidence in terms of label distribution learning. Although the current approach has limitations in that the method for generating labels is relatively simple and limited to image classification, through experiments and visualization on four data sets, the LDL-based training approach can simultaneously improve model accuracy and calibration. As a future study, we plan to verify the utility of LDL in other tasks and analyze LDL based on theoretical backgrounds such as class-considered approaches and risk minimization [41].

\section*{References}
{
\small
[1] Krizhevsky, A., Sutskever, I.\ \& Hinton, G.E.\ (2012) Imagenet classification with deep convolutional neural networks. {\it In: Proc. of NeurIPS}.

[2] Simonyan, K.\ \& Zisserman, A.\ (2015) Very deep convolutional networks for large-scale image recognition. {\it In: Proc. of ICLR}.

[3] He, K., Zhang, X., Ren, S.\ \& Sun, J.\ (2016) Deep residual learning for image recognition. {\it In: Proc. of CVPR}.

[4] Guo, C., Pleiss, G., Sun, Y.\ \& Weinberger, K.Q.\ (2017) On calibration of modern neural networks. {\it In: Proc. of ICML}.

[5] Thulasidasan, S., Chennupati, G., Bilmes, J.A., Bhattacharya, T.\ \& Michalak, S.E.\ (2019) On mixup training: Improved calibration and predictive uncertainty for deep neural networks. {\it In: Proc. of NIPS}.

[6] Maronas, J., Ramos-Castro, D.\ \& Palacios, R.P.\ (2020) Improving calibration in mixup-trained deep neural networks through confidence-based loss functions. {\it arXiv:2003.09946}. 

[7] Hinton, G., Vinyals, O.\ \& Dean, J.\ (2015) Distilling the Knowledge in a Neural Network. {\it In: Proc. of NeurIPS Workshop}.

[8] Romero, A., Ballas, N., Kahou, S.E., Chassang, A., Gatta, C.\ \& Bengio, Y.\ (2015) Fitnets: Hints for thin deep nets. {\it In: Proc. of ICLR}.

[9] Furlanello, T., Lipton, Z., C., Tschannen, M., Itti, L.\ \& Anandkumar, A.\ (2018) Born again neural networks. {\it In: Proc. of ICML}.

[10] Park, W., Kim, D., Lu, Y.\ \& Cho, M.: Relational knowledge distillation.\ (2019) {\it In: Proc. of CVPR}.

[11] Shen, Z.\ \& Savvides, M.\ (2020) Meal v2: Boosting vanilla resnet-50 to 80\%+ top-1 accuracy on imagenet without tricks. {\it arXiv:2009.08453}.

[12] Tian, Y., Krishnan, D.\ \& Isola, P.\ (2020) Contrastive representation distillation. {\it In: Proc. of ICLR}.

[13] Wu, G.\ \& Gong, S.\ (2020) Peer collaborative learning for online knowledge distillation. {\it In: Proc. of CVPR}.

[14] Yun, S., Park, J., Lee, K.\ \& Shin, J.\ (2020) Regularizing class-wise predictions via selfknowledge distillation. {\it In: Proc. of CVPR}.

[15] Xu, G., Liu, Z., Li, X.\ \& Loy, C.C.\ (2020) Knowledge distillation meets self-supervision. {\it In: Proc. of ECCV}.

[16] Zhou, H., Song, L., Chen, J., Zhou, Y., Wang, G., Yuan, J.\ \& Zhang, Q.\ (2021) Rethinking soft labels for knowledge distillation: A bias-variance tradeoff perspective. {\it In: Proc. of ICLR}.

[17] Szegedy, C., Vanhoucke, V., Ioffe, S.\ \& Shlens, J.\ (2016) Rethinking the inception architecture for computer vision. {\it In: Proc. of CVPR}.

[18] Muller, R., Kornblith, S.\ \& Hinton, G.\ (2019) When does label smoothing help? {\it In: Proc. of NeurIPS}.

[19] Shen, Z., Liu, Z., Xu, D., Chen, Z., Cheng, K.T.\ \& Savvides, M.\ (2021) Is label smoothing truly incompatible with knowledge distillation: An empirical study. {\it In: Proc. of ICLR}.

[20] Geng, X., Smith-Miles, K.\ \& Zhou, Z.H.\ (2010) Facial age estimation by learning from label distributions. {\it In: Proc. of AAAI}.

[21] Gao, B.B., Xing, C., Xie, C.W., Wu, J.\ \& Geng, X.\ (2017) Deep label distribution learning with label ambiguity. {\it IEEE Transactions on Image Processing}, {\bf 26}(6).

[22] Xu, N., Qiao, C., Geng, X.\ \& Zhang, M.L.\ (2021) Instance-dependent partial label learning. {\it In: Proc. of NeurIPS}.

[23] Krizhevsky, A.\ (2009) Learning multiple layers of features from tiny images. {\it In: Technical Report}.

[24] Coates, A., Ng, A.\ \& Lee, H.\ (2011) An analysis of single-layer networks in unsupervised feature learning. {\it In: Proc. of AISTAT}.

[25] Deng, J., Dong, W., Socher, R., Li, L.J., Li, K.\ \& Fei-Fei, L.\ (2009) Imagenet: A large-scale hierarchical image database. {\it In: Proc. of CVPR}.

[26]  Zagoruyko, S.\ \& Komodakis, N.\ (2017) Paying more attention to attention: Improving the performance of convolutional neural networks via attention transfer. {\it In: Proc. of ICLR}.

[27] Yim, J., Joo, D., Bae, J.\ \& Kim, J.\  (2017) A gift from knowledge distillation: Fast optimization, network minimization and transfer learning. {\it In: Proc. of CVPR}.

[28] Zhou, H., Song, L., Chen, J., Zhou, Y., Wang, G., Yuan, J.\ (2021) \& Zhang, Q.\ Rethinking soft labels for knowledge distillation: A bias-variance tradeoff perspective. {\it In: Proc. of ICLR}.

[29] Yuan, L., Tay, F.E., Li, G., Wang, T.\ \& Feng, J.\ (2020) Revisiting knowledge distillation via label smoothing regularization. {\it In: Proc. of CVPR}.

[30] Zhang, Z., Dalca, A.V.\ \& Sabuncu, M.R.\ (2019) Confidence calibration for convolutional neural networks using structured dropout. {\it arXiv:1906.09551}.

[31] Seo, S., Seo, P.H.\ \& Han, B.\ (2019) Learning for single-shot confidence calibration in deep neural networks through stochastic inferences. {\it In: Proc. of CVPR}.

[32] Krishnan, R.\ \& Tickoo, O.\ (2020) Improving model calibration with accuracy versus uncertainty optimization. {\it In: Proc. of NeurIPS}.

[33] Zhang, H., Cisse, M., Dauphin, Y.N.\ \& Lopez-Paz, D.\ (2018) mixup: Beyond empirical risk minimization. {\it In: Proc. of ICLR}.

[34] Takahashi, R., Matsubara, T.\ \& Uehara, K.\ (2019) Data augmentation using random image cropping and patching for deep cnns. {\it IEEE Transactions on Circuits and Systems for Video Technology}, {\bf 30}(9).

[35] Yun, S., Han, D., Oh, S.J., Chun, S., Choe, J.\ \& Yoo, Y.\ (2019) Cutmix: Regularization strategy to train strong classifiers with localizable features. {\it In: Proc. of ICCV}.

[36] Xie, S., Girshick, R., Doll´ar, P., Tu, Z.\ \& He, K.\ (2017) Aggregated residual transformations for deep neural networks. {\it In: Proc. of CVPR}.

[37] Huang, G., Liu, Z., van der Maaten, L.\ \& Weinberger, K., Q.\ (2017) Densely connected convolutional networks. {\it In: Proc. of CVPR}.

[38] Yu, F., Wang D., Shelhamer, E.\ \& Darrell, T.\ (2018) Deep layer aggregation. {\it In: Proc. of CVPR}.

[39] DeVries, T.\ \& Taylor, G.W.\ (2017) Improved regularization of convolutional neural networks with cutout. {\it arXiv:1708.04552}.

[40] Laves, M.-H., Ihler, S., Kortmann, K.-P.,\ \& Ortmaier, T.\ (2019) Well-calibrated model uncertainty with temperature scaling for dropout variational inference. {\it NeurIPS Workshop}.

[41] Balestriero, R., Bottou, L.,\ \& LeCun, Y.\ (2022) The effects of regularization and data augmentation are class dependent. {\it arXiv:2204.03632}.

[42] Gal, Y.\ \& Ghahramani, Z.\ (2016) Dropout as a Bayesian approximation: Representing model uncertainty in deep learning. {\it In: Proc. of ICML}.

[43] Wang, W., Tran, D.\ \& F. Matt.\ (2020) What makes training multi-modal classification networks hard? {\it In: Proc. of CVPR}.
}

\section*{Supplementary Materials}

\section*{Overview}
This supplementary material contains additional experiments and analyses not included in the main manuscript. 1) Experimental results with uncertainty based LDL, 2) Additional model calibration error measurement results, 3) Loss curve and training tendency, 4) Additional visualization of feature representation, 5) LDL evaluation results for each dataset according to the student/teacher configurations, 6) Pseudocode of LDL, and 7) Reliability diagrams according to each method are included.

\section{Uncertainty based LDL}

In addition to data augmentation, we evaluated uncertainty-based LDL as an online LDL method for generating labels during training. Uncertainty-based LDL adopts variational inference [42] on teacher networks to generate random labels $\hat{z}=\frac{1}{N}\sum_{i=1}^{N}\sigma(\theta_{t,i}(x))$, where $N$ is the total number of teacher networks dropped from MCDO and $\sigma$ is the softmax function. We set the drop parameters of MCDO $\rho$ to 0.2 and N to 20 through empirical experiment. We used the mean of the softmax output for MC dropout-based variance inference as $\hat{z}$.

We have applied another method for uncertainty-based LDL. Overfitting-to-generalization-ratio (OGR) is originally designed to prevent overfitting during mutlmodal training by defining the tendency of the gradient of each modality during training as an adaptive loss [43]. We modified the OGR as per class loss difference between validation and training loss and use it as an adaptive loss or as label distribution by normalized weighting one-hot label. The OGR is defined as $OGR_k=\frac{((L_{val,N+n}-L_{train,N+n})-(L_{val,N}-L_{train_N}))^2}{L_{val,N+n}-L_{val,N}}$,  where given model parameters $\theta$ at epoch $N$, $L$ is average loss of each $k$th class over the fixed train set, and $n$ is gap of epoch. We set $n$ to 10. 

Table 1 shows the results obtained with each approach. Uncertainty-based LDL methods were also able to achieve mostly improved performance compared to vanilla models, but the improvement was small compared to on/offline LDL methods. Nevertheless, in almost all configurations of MCDO, a positive effect could also be achieved in model calibration. In the case of OGR, the gain was not even greater than that of MCDO, but when OGR was applied to the LDL method, more improved results were obtained in accuracy and model correction than when the class-wise loss was used. It was confirmed that overconfidence occurs when OGR is used as a class-wise loss. We speculate that the reason why MCDO and OGR are difficult to achieve as much performance improvement as the online LDL technique is that they generate the label distribution through only the change of the label without considering the data. Nevertheless, uncertainty-based LDL was still able to validate its potential in accuracy and model calibration. 

\begin{table}[h!]
\caption{Classification accuracy and ECE of the uncertainty based LDL and other approaches for the CIFAR 100. MCDO [42] based LDL are marked with T and S according to the application of the teacher and student networks. OGR [42] includes two approaches: each class OGR is reflected in loss or LDL is reflected.}
\begin{center}
\label{uncertainty}
\scalebox{0.75}{
\begin{tabular}{cccc}
\hline
Teacher 					& 							& ResNet200					& 				\\ 
Student					& ResNet18					& ResNet50					& ResNet200		\\ \hline
Vanilla 					& 77.83±0.55/0.080				& 78.57±0.35/0.107				& 79.47±0.58/0.101 \\ 
Temperature Scaling [4]		& 77.83±0.55/0.039				& 78.57±0.35/0.032				& 79.47±0.58/\textbf{0.029} \\ 
Label smoothing			& 78.95±0.06/0.085				& 78.90±0.24/0.042				& 79.59±0.42/0.037 \\ \hline
KD ($\alpha$=0.1,$T$=3) [7]	& 77.73±0.16/0.077				& 79.12±0.45/0.103				& 80.10±0.23/0.100 \\ 
CRD+KD [12]				& 80.41±0.14/0.108				& 80.34±0.07/0.118				& 81.58±0.20/0.110 \\ 
WSL	[16]			 		& 79.91±0.03/0.111 				& 80.25±0.12/0.114				& 80.96±0.01/0.107 \\ \hline
MCDO [42]				& 78.02±0.03/0.010 				& 79.38±0.14/0.104				& 79.80±0.23/0.110 \\ 
On-LDL (MCDO-T)			& 77.95±0.13/0.072 				& 78.26±0.98/0.092				& 79.56±0.45/0.084 \\ 
On-LDL (MCDO-S, T)		& 78.06±0.34/0.091				& 79.23±0.18/0.089				& 80.64±0.12/0.077 \\ 
On-LDL (MCDO-T, CutMix)	& 80.82±0.04/0.059				& 82.26±0.25/0.032				& 83.61±0.11/0.047 \\ 
OGR [43]					& 77.79±0.15 / 0.370 			& 78.64±0.17 / 0.352				& 79.18±0.18 / 0.352 \\ 
On-LDL (OGR)				& 78.47±0.27 / 0.077 			& 79.58±0.24 / 0.101				& 79.36±0.32 / 0.107 \\ \hline
On-LDL					& 81.26±0.11/0.043				& 83.09±0.05/\textbf{0.030}		& 83.88±0.28/0.038 \\ 
On-LDL+					& \textbf{81.76±0.25}/\textbf{0.034}	& \textbf{83.57±0.05}/0.038		& \textbf{84.44±0.13}/0.039 \\ \hline
\end{tabular}
}
\end{center}
\end{table}

\section{Additional Model Calibration Error Measurement}
In addition to the ECE in the experiments, different calibration error metrics include uncertainty calibration error (UCE), negative log-likelihood (NLL), and Brier score [6, 7, 8]. Table 2 shows that for most model calibration error measurement metrics, the LDL-based method achieves the lowest error except for temperature scaling, which is simple normalization. In particular, the NLL and Brier scores achieve the lowest error in all cases. In the case of MCDO-based LDL, improved results were observed in ECE, but high errors were observed in the remaining metrics, and very high in NLL and Brier.

\begin{table}[h!]
\caption{Evaluation results according to model calibration error scores (ECE, UCE, NLL, and Brier) with different teacher (T) and student (S) configuration in CIFAR 100.}
\begin{center}
\label{uncertainty}
\scalebox{0.75}{
\begin{tabular}{cccccc}
\hline
T(R200)/S(R18)			& 					&			& 				& 			&			\\ 
Score Metric				& acc				&	 ece		&	uce			& nll			& Brier		\\ \hline
Vanilla 					& 77.83±0.55			&	0.080	&	0.106		& 0.914		& 0.125		\\ 
Temperature Scaling [4]		& 77.83±0.55			&	0.039	&	\textbf{0.033}	& 0.892		& 0.121		\\ 
Label smoothing			& 78.95±0.06			&	0.085	& 	0.138		& 0.964		& 0.122		\\  \hline
KD ($\alpha$=0.1,$T$=3) [7]	& 77.73±0.16			&	0.077	& 	0.104		& 0.905		& 0.124		\\ 
CRD+KD [12]				& 80.41±0.14			&	0.108	& 	0.135		& 0.903		& 0.128		\\ 
WSL	[16]			 		& 79.91±0.03			&	0.111 	& 	0.137		& 0.936		& 0.131		\\  \hline
Off-LDL					& 78.67±0.14			&	0.060 	& 	0.086		& 0.826		& 0.118		\\ 
On-LDL (MCDO-S, T)		& 78.06±0.34			&	0.091	&	0.123		& 1.918		& 0.336		\\ 
On-LDL					& 81.26±0.11			&	0.043	&	0.078		& 0.801		& 0.112		\\ 
On-LDL+					& \textbf{81.76±0.25}		&\textbf{0.034}	&	0.073		&\textbf{0.693}	& \textbf{0.103}	\\  \hline \hline

T(R200)/S(R50)			& 					&			& 				& 			&			\\ 
Score Metric				& acc				&	 ece		&	uce			& nll			& Brier		\\ \hline
Vanilla 					& 78.57±0.35			&	0.107	&	0.137 		& 0.933		& 0.131		\\ 
Temperature Scaling [4]		& 78.57±0.35			&	0.032	&	0.044		& 0.815		& 0.114		\\ 
Label smoothing			& 78.90±0.24			&	0.042	& 	0.085		& 0.939		& 0.116		\\  \hline
KD ($\alpha$=0.1,$T$=3) [7]	& 79.12±0.45			&	0.103	& 	0.133 		& 0.914		& 0.129		\\ 
CRD+KD [12]				& 80.34±0.07			&	0.118	& 	0.143		& 0.943		& 0.132		\\ 
WSL	[16]			 		& 80.25±0.12			&	0.114 	& 	0.139		& 0.938		& 0.129		\\  \hline
Off-LDL					& 79.19±0.37  			&	0.087 	& 	0.116 		& 0.835		& 0.122		\\ 
On-LDL (MCDO-S, T)		& 79.23±0.18			&	0.089	&	0.120 		& 2.871		& 0.543		\\ 
On-LDL					& 83.09±0.05			&\textbf{0.030}	&\textbf{0.064}		& 0.633		& \textbf{0.095}	\\ 
On-LDL+					& \textbf{83.57±0.05}		&	0.038	&	0.065 		&\textbf{0.621}	& 0.096	\\  \hline \hline

T(R200)/S(R200)			& 					&			& 				& 			&			\\ 
Score Metric				& acc				&	 ece		&	uce			& nll			& Brier		\\ \hline
Vanilla 					& 79.47±0.58			&	0.101	&	0.128 		& 0.895		& 0.126		\\ 
Temperature Scaling [4]		& 79.47±0.58			&\textbf{0.029}	&	0.033		& 0.782		& 0.111		\\ 
Label smoothing			& 79.59±0.42			&	0.037	& 	0.073		& 0.907		& 0.111		\\  \hline
KD ($\alpha$=0.1,$T$=3) [7]	& 80.10±0.23			&	0.100	& 	0.128 		& 0.873		& 0.124		\\ 
CRD+KD [12]				& 81.58±0.20			&	0.110 	& 	0.134		& 0.879		& 0.124		\\ 
WSL	[16]			 		& 80.96±0.01			&	0.081	& 	0.109		& 0.787		& 0.125		\\  \hline
Off-LDL					& 80.60±0.30  			&	0.087 	& 	0.116 		& 0.835		& 0.122		\\ 
On-LDL (MCDO-S, T)		& 80.64±0.12			&	0.077	&	0.106 		& 3.848		& 0.766		\\ 
On-LDL					& 83.88±0.28			&	0.038 	&	0.067		& 0.607		& 0.094		\\ 
On-LDL+					& \textbf{84.44±0.13}		&	0.039 	&	\textbf{0.065} 	& \textbf{0.591}	& \textbf{0.092}	\\  \hline \hline
\end{tabular}
}
\end{center}
\end{table}

\section{Loss Curve and Training Tendency}
\begin{figure*}[h!]
\begin{center}
\includegraphics[width=0.7\linewidth]{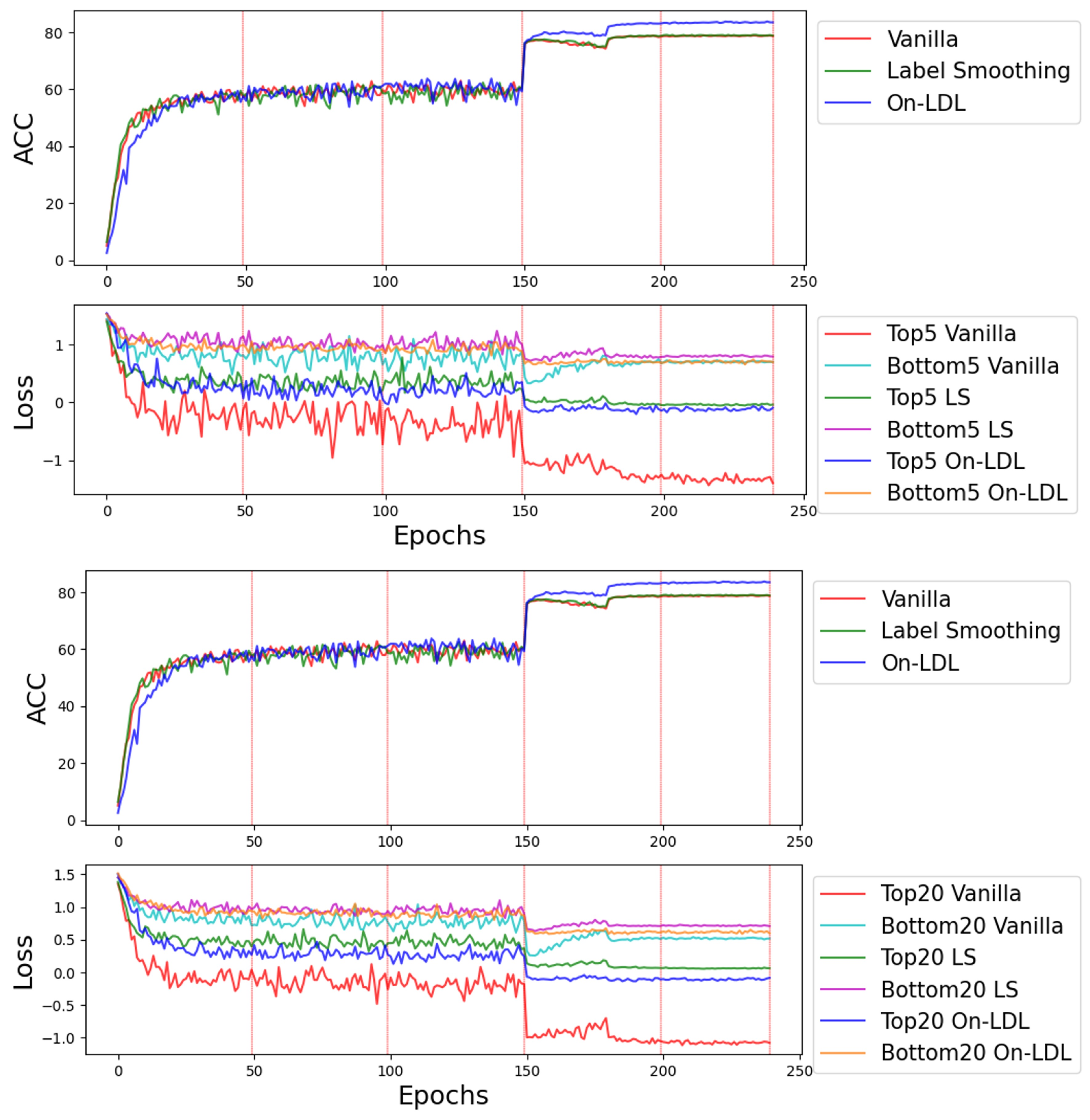}
\caption{\textbf{Log loss and accuracy plots (ResNet50) for top and bottom (5 and 20 classes) for CIFAR100.} The upper graph plotted the top and bottom 5 classes, and the bottom graph plotted 20 classes.} 
\label{log_loss}
\end{center}
\end{figure*}

Figure \ref{log_loss} shows the loss curves according to ResNet50-based vanilla, label smoothing, and online LDL methods in CIFAR100 and training trends according to accuracy. The loss curve plotted the log loss of the top 5 and bottom 5 accuracy classes to investigate the training trends for each class.
The most notable feature in the trend of the loss curve is the largest difference between the top 5 and bottom 5 losses for vanilla training. LDL and label smoothing suppress this tendency, and the tendency becomes more pronounced after the first drop in learning rate. LDL has the worst test performance at the beginning of training but has the highest accuracy after the first drop in learning rate. This tendency is also observed when the log loss for the top and bottom 20 classes is plotted. We observed the effect of label distribution learning in preventing overconfidence by class, and then we are considering an approach that can adaptively apply label distribution and data augmentation by class. The trend of this loss curve is similarly observed in previous studies [9, 10], but we were able to observe the cause more clearly in the log loss by class, and the online LDL makes a more firm contribution to the generalization performance.

\clearpage

\section{Additional Visualization of Feature Representation}
\begin{figure*}[h!]
\begin{center}
\includegraphics[width=1.0\linewidth]{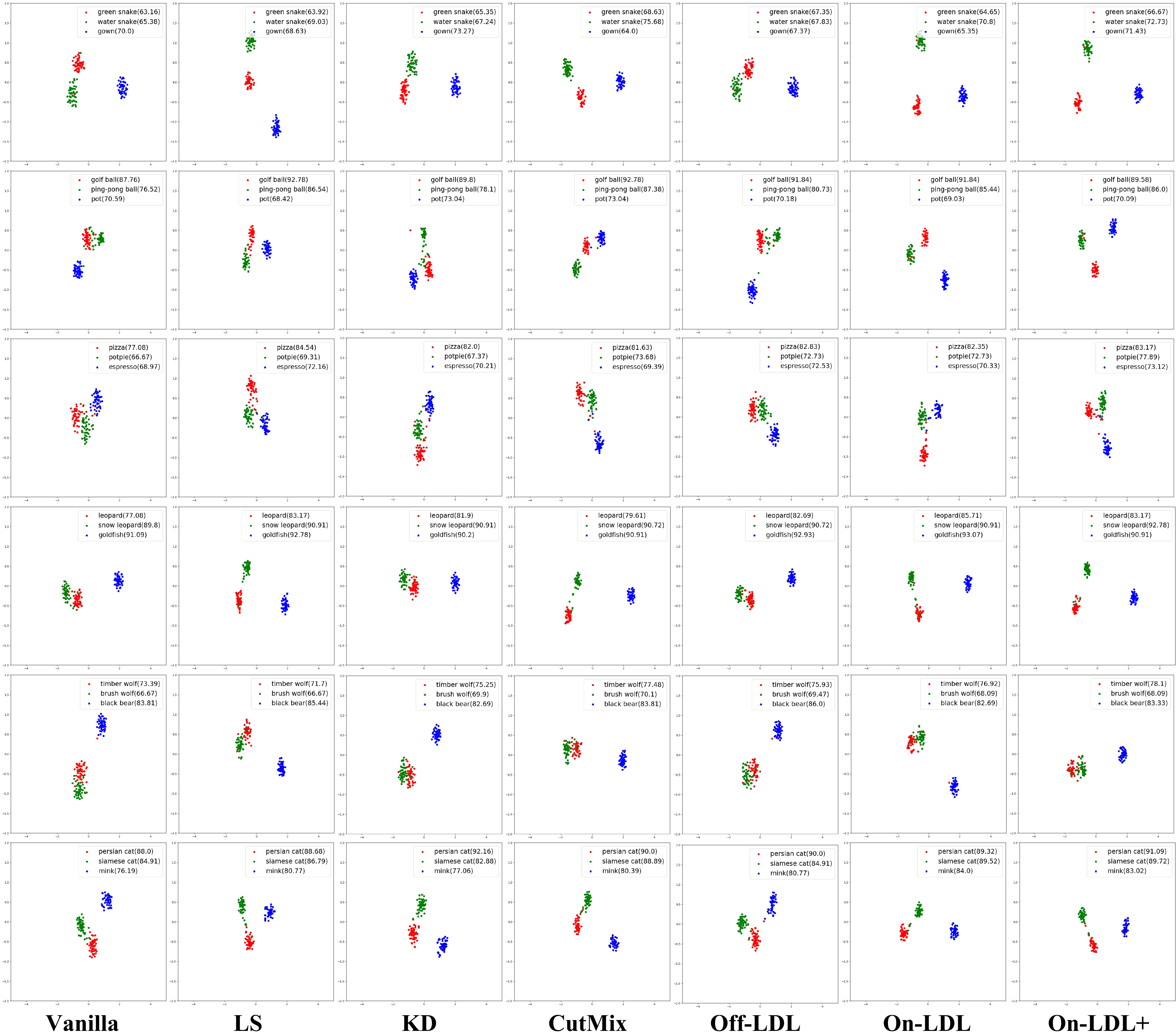}
\caption{\textbf{Visualization of penultimate layer's activation of ResNet50 for ImageNet.} From the top: ['green snake','water snake','grown'], ['golf ball','ping-pong ball','pot'], ['pizza','potpie','espresso'], ['leopard','snow leopard','goldfish'], ['timber wolf','brush wolf','black bear'], ['persian cat','siamese cat','mink ']. The first two classes are semantically similar, and the last class is a semantically different class.} 
\label{feature_vis}
\end{center}
\end{figure*}

Figure \ref{feature_vis} shows additional visualization of the activation of the penultimate layer of ResNet50 for ImageNet dataset. Each plot consists of two semantically similar classes of ImageNet and one semantically different class. From the first line in Figure \ref{feature_vis}: ['green snake','water snake','grown'], ['golf ball','ping-pong ball','pot'], ['pizza','potpie','espresso'], ['leopard','snow leopard','goldfish'], ['timber wolf','brush wolf','black bear'], ['persian cat','siamese cat','mink ']. The first two classes are semantically similar, and the last class is a semantically different class.  Through penultimate layer visualization, it is observed that the on/offline LDL technique generates an effective representation for classifying semantically similar classes except the case of ['timber wolf','brush wolf','black bear'] (5th row).

\clearpage

\section{LDL Model Comparison}
\begin{figure*}[h!]
\begin{center}
\includegraphics[width=1.0\linewidth]{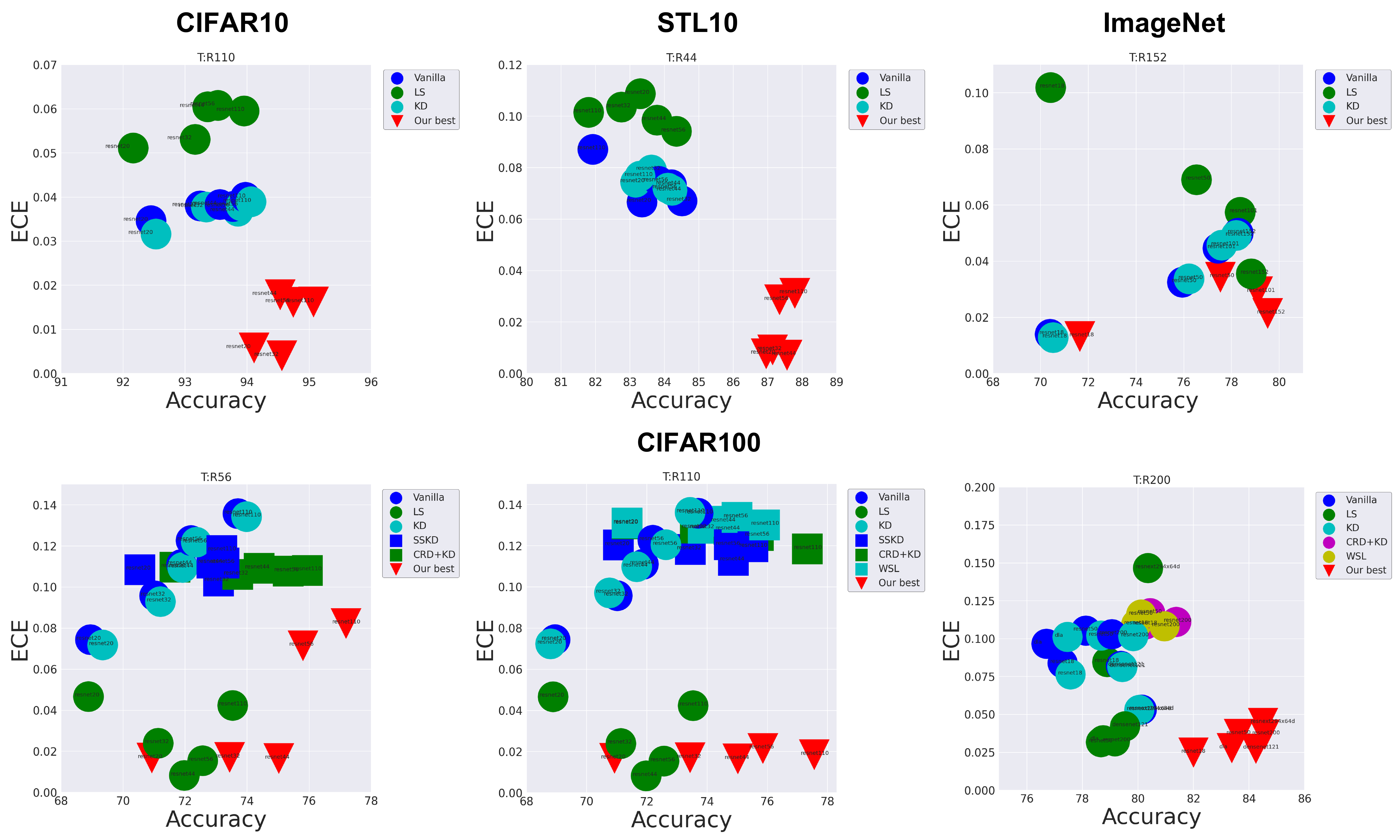}
\caption{\textbf{Comparison of performance across specific datasets and teacher networks.} The x-axis is classification accuracy, and the y-axis is ECE. Each graph was plotted in different colors and shapes according to the training methodology, and each model name is indicated in each figure.} 
\label{model_comparisions}
\end{center}
\end{figure*}

Figure \ref{model_comparisions} shows a comparison plot between the proposed LDL-based best-performing model and other training methods. For each plot, the x-axis is classification accuracy and the y-axis is ECE. Each graph shows the accuracy/ECE relationship for all student networks trained from a specific teacher network. The CNN models with the LDL approach in all datasets are located in the lower right of the graph, which means that the LDL achieves low ECE and high classification accuracy.

\clearpage

\section{Implementation of On/offline LDL}
\begin{lstlisting}[language=Python]
import torch
import torch.nn as nn
import numpy as np

criterion = nn.CrossEntropyLoss()
for images, labels in loader:
    # Generate data augments Images
    images = Data_Aug(image, aug_method)
    
    # Generate soft labels
    if method in ['Off-LDL', 'On-LDL']:
        with torch.no_grad():
            labels = TeacherModel(images)
   
    logits = StudentModel(images)
    
    # Cross entropy loss
    loss = criterion(logits, labels)
    
def Data_Aug(images, aug_method, alpha=1.0):
    lam = np.random.beta(alpha, alpha)
    I_x, I_y = images.size()[2:]
    # shuffle minibatch
    index = torch.randperm(images.size(0))
    rand_image = images[index]
    if aug_method == 'mixup': # MixUp Algorithm
        images = lam * images + (1 - lam) * rand_image
    elif aug_method == 'ricap': #RICAP Algorithm
        # draw a boundary position (w,h)
        w = int(np.round(I_x * np.random.beta(alpha, alpha)))
        h = int(np.round(I_y * np.random.beta(alpha, alpha)))
        w_ = [w, I_x-w, w, I_x-w]
        h_ = [h, I_y-h, h, I_y-h]
        # select and crop four images
        cropped_images = {}
        for k in range(4):
            index = torch.randperm(images.size(0))
            x_k = np.random.randint(0, I_x - w_[k] + 1)
            y_k = np.random.randint(0, I_y - h_[k] + 1)
            cropped_images[k] = images[index][:, :, x_k:x_k + w_[k], y_k:y_k + h_[k]]
        images = torch.cat(
            (torch.cat((cropped_images[0], cropped_images[1]), 2)
            torch.cat((cropped_images[2], cropped_images[3]), 2)), 3)
    elif aug_method == 'cutmix': # CutMix Algorithm
        cut_w = np.int(I_x * np.sqrt(1. - lam)) # cut rate
        cut_h = np.int(I_y * np.sqrt(1. - lam))
    
        cx = np.random.randint(I_x)
        cy = np.random.randint(I_y)
    
        bbx1 = np.clip(cx - cut_w // 2, 0, I_x)
        bby1 = np.clip(cy - cut_h // 2, 0, I_y)
        bbx2 = np.clip(cx + cut_w // 2, 0, I_x)
        bby2 = np.clip(cy + cut_h // 2, 0, I_y)
        images[:, :, bbx1:bbx2, bby1:bby2] = rand_image[:, :, bbx1:bbx2, bby1:bby2]
        
    return images
\end{lstlisting} 

\section{Reliability Diagram}
\begin{figure*}[h!]
\begin{center}
\includegraphics[width=1.0\linewidth]{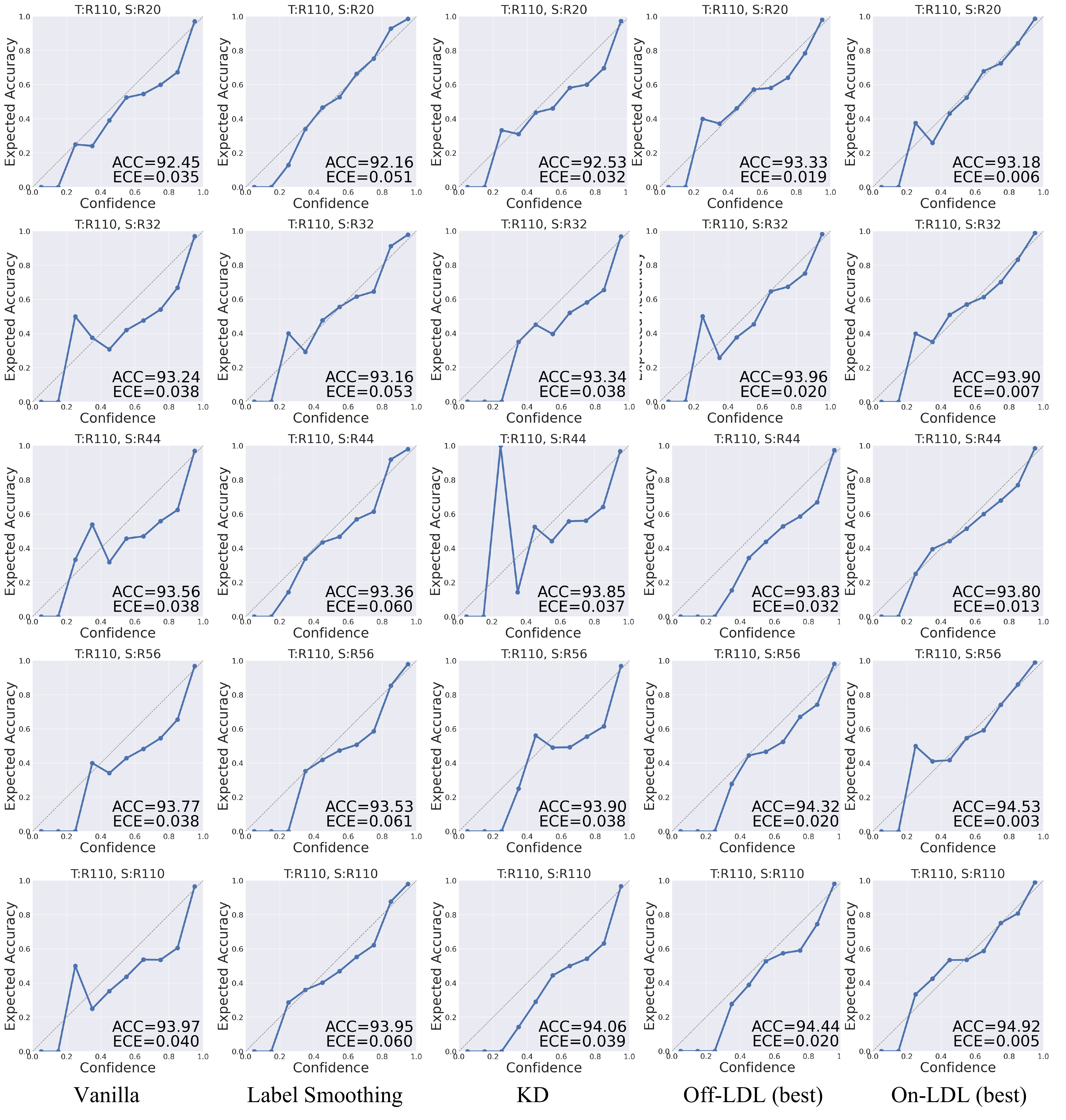}
\caption{\textbf{Reliability diagram for CIFAR10.} Accuracy and ECE for the dataset are indicated for each method.} 
\label{rd_cifar10}
\end{center}
\end{figure*}

\begin{figure*}[h!]
\begin{center}
\includegraphics[width=1.0\linewidth]{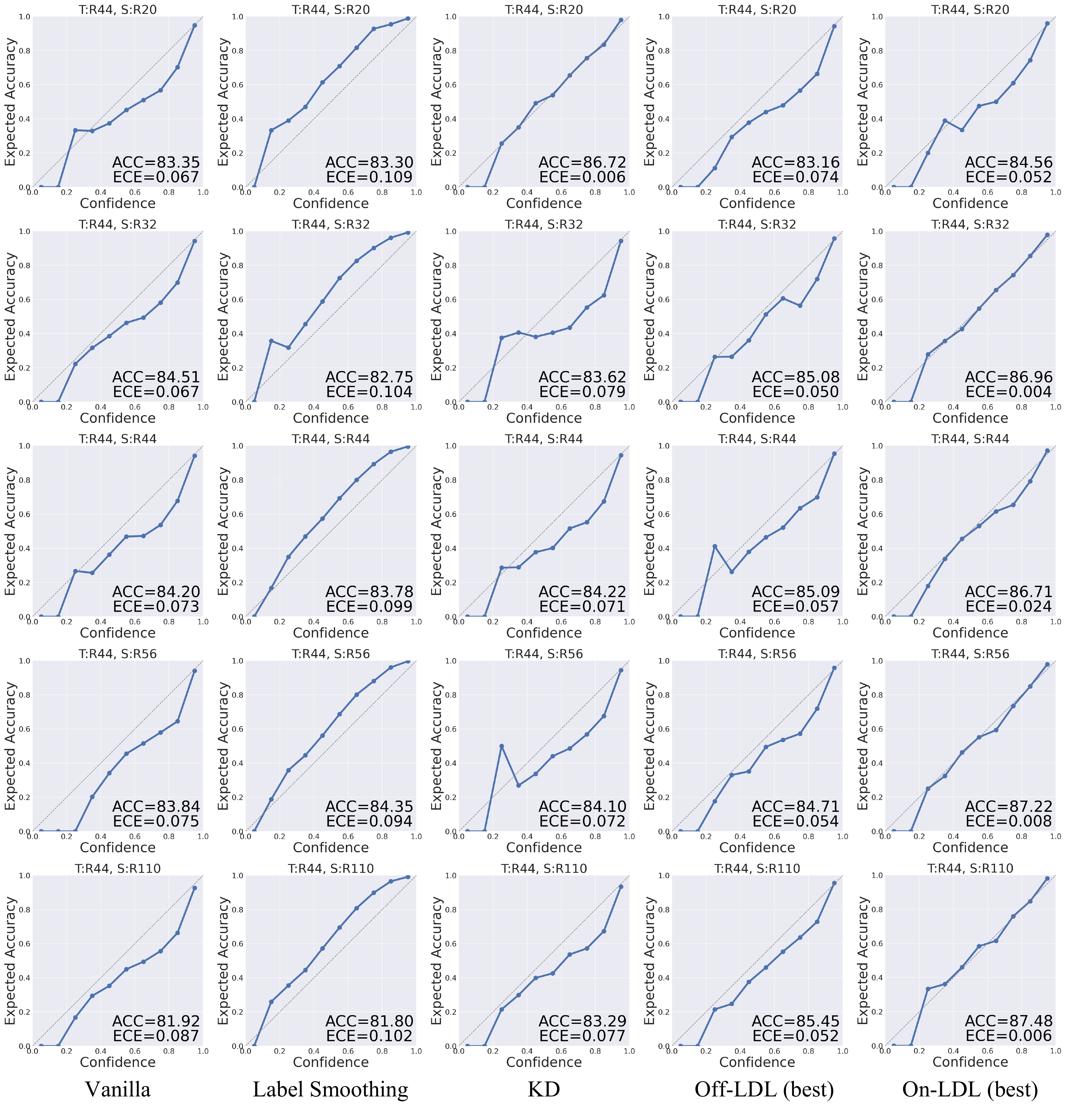}
\caption{\textbf{Reliability diagram for SLT10.} Accuracy and ECE for the dataset are indicated for each method.} 
\label{rd_stl10}
\end{center}
\end{figure*}

\begin{figure*}[h!]
\begin{center}
\includegraphics[width=1.0\linewidth]{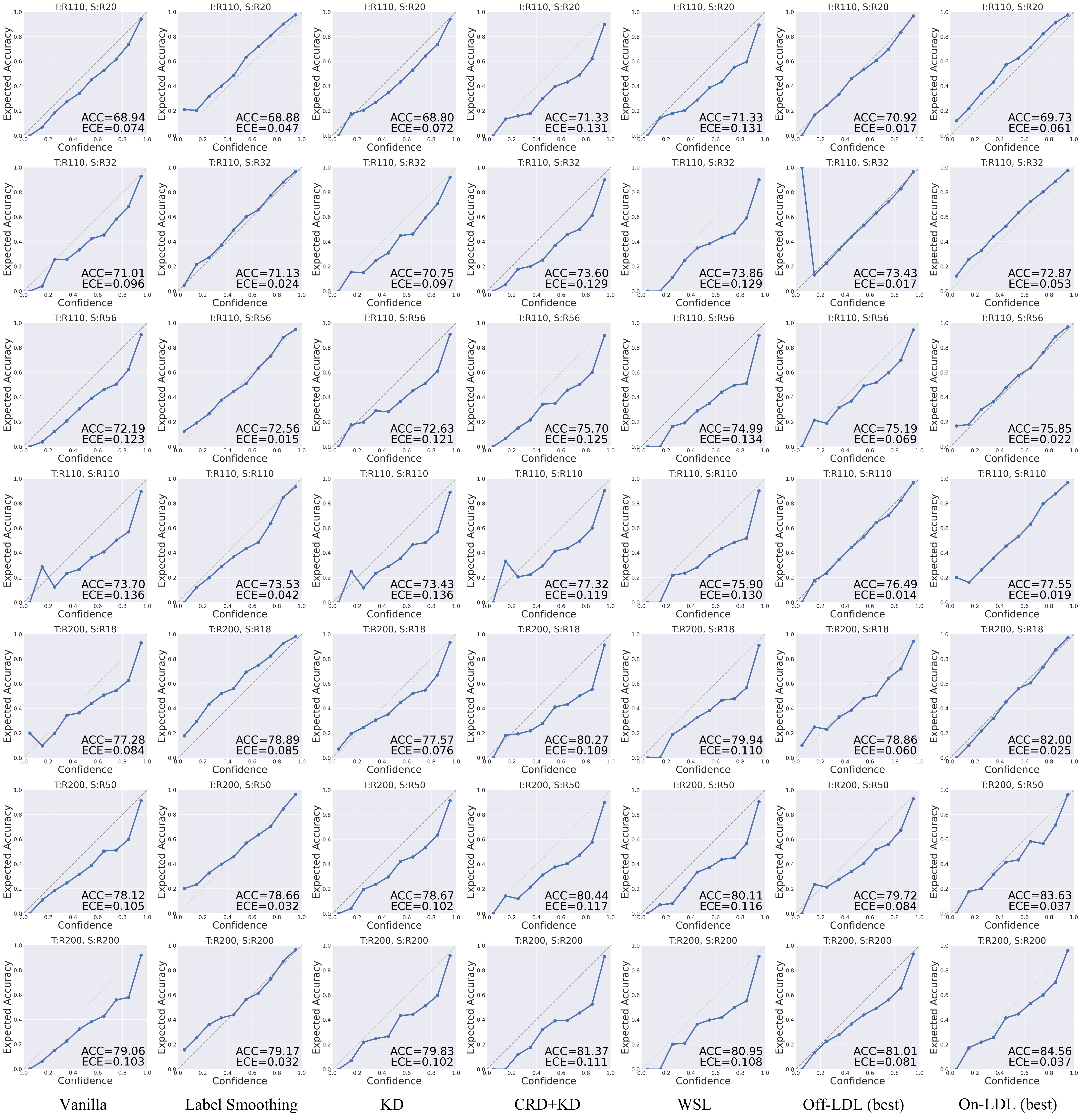}
\caption{\textbf{Reliability diagram for CIFAR100.} Accuracy and ECE for the dataset are indicated for each method.} 
\label{rd_cifar100}
\end{center}
\end{figure*}

\begin{figure*}[h!]
\begin{center}
\includegraphics[width=1.0\linewidth]{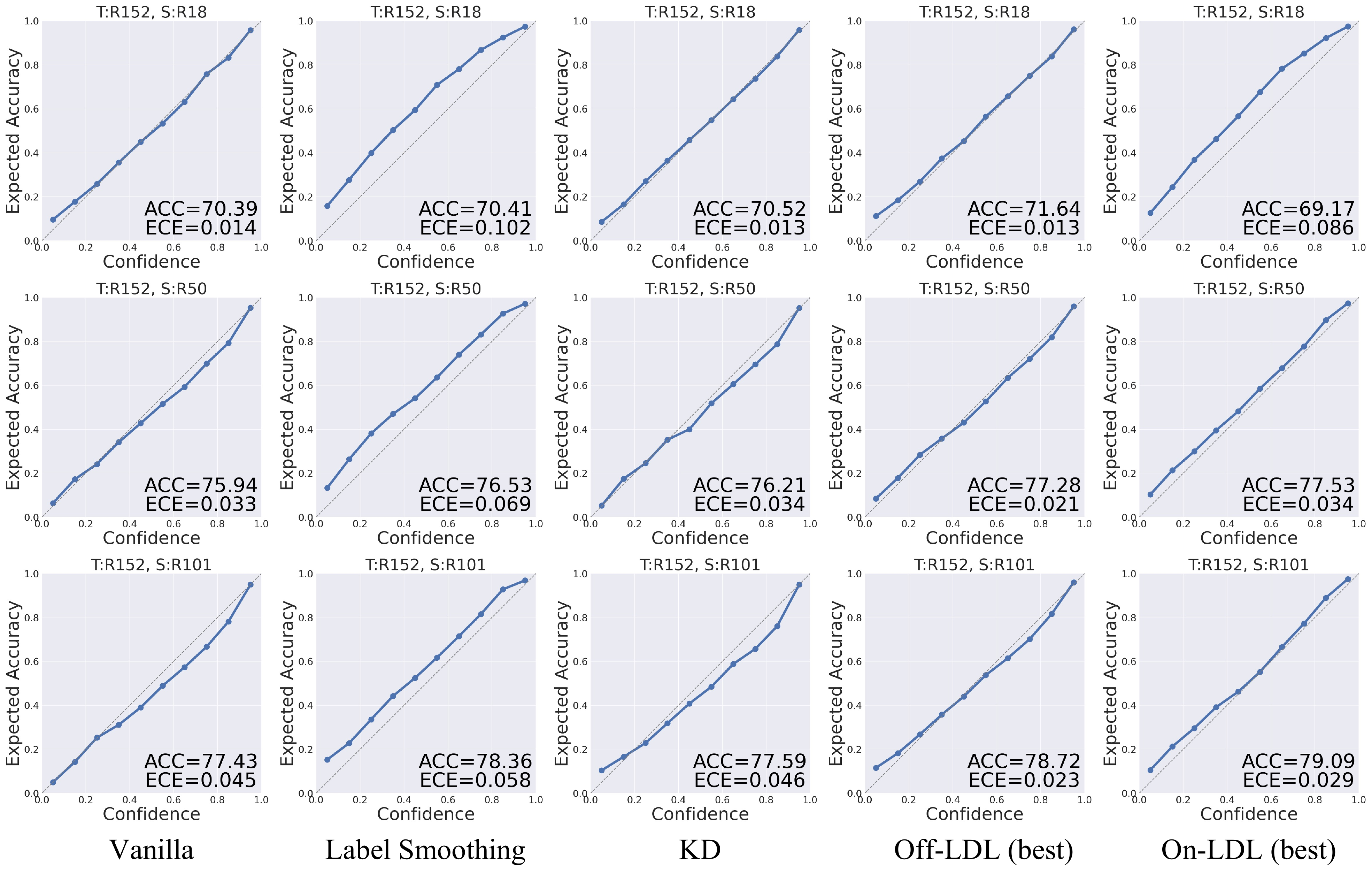}
\caption{\textbf{Reliability diagram for ImageNet.} Accuracy and ECE for the dataset are indicated for each method.} 
\label{rd_imagenet}
\end{center}
\end{figure*}

Reliability diagram can intuitively visualize the correlation between model confidence and inference accuracy based on a histogram [6]. Figure \ref{rd_cifar10}, \ref{rd_stl10}, \ref{rd_cifar100}, and \ref{rd_imagenet} show the reliability diagram for each dataset and approach. The x-axis of the reliability diagram is a histogram of the confidence at a specific interval, and the y-axis is the expected value of the accuracy for the inferred confidence. The better the model correction effect is, the more the plot values match the linear lines for the x and y axes [6].

\end{document}